%% file: main.tex
\documentclass[11pt]{article}
\usepackage{coling2020}
\usepackage{times}
\usepackage{url}
\usepackage{latexsym}
\colingfinalcopy

\usepackage{amsmath}
\usepackage{amssymb}
\usepackage{amsfonts}

\usepackage{enumerate}
\usepackage{enumitem}

\usepackage{algorithm}
\usepackage{algpseudocode}

\usepackage{graphicx}
\usepackage{tabularx}
\usepackage{soul}

\usepackage[nodisplayskipstretch]{setspace}
\usepackage{hyperref}
\usepackage{xstring}

\usepackage{tikz}
\usepackage{psfrag}
\usepackage{colortbl}

\usepackage{wrapfig}

\usepackage{subcaption}

\usetikzlibrary{bayesnet}
\usetikzlibrary{dsp}

\allowdisplaybreaks
\newcommand{\norm}[1]{\left\lVert#1\right\rVert}

\title{Variational Autoencoder with Embedded Student-$t$ Mixture Model \\ for Authorship Attribution}

\author{Benedikt Boenninghoff,$^1$ Steffen Zeiler,$^1$ Robert M. Nickel,$^2$ Dorothea Kolossa$^1$ \\
  $^1$\textit{Cognitive Signal Processing Group, Ruhr University Bochum, Germany} \\
  $^2$\textit{Department of Electrical and Computer Engineering, Bucknell University, Lewisburg, PA, USA}\\
  {\tt \{benedikt.boenninghoff, dorothea.kolossa, steffen.zeiler\}@rub.de} \\
   {\tt rmn009@bucknell.edu} \\
  }
\date{}

\usepackage[acronym]{glossaries}
\glsdisablehyper

\newacronym{VAE}{VAE}{Variational Autoencoder}
\newacronym{GMM}{GMM}{Gaussian Mixture Models}
\newacronym{SMM}{SMM}{Student-$t$ Mixture Model}
\newacronym{i.i.d.}{i.i.d.}{independent and identically distributed}
\newacronym{ELBO}{ELBO}{Evidence Lower Bound}
\newacronym{tVAE}{tVAE}{SMM-based Variational Autoencoder}
\newacronym{gVAE}{gVAE}{GMM-based Variational Autoencoder}
\newacronym{KL}{KL}{Kullback-Leibler}
\newacronym{AE}{AE}{Autoencoder}
\newacronym{SVM}{SVM}{Support Vector Machine}

\begin{document}

\maketitle

\begin{abstract}
Traditional computational authorship attribution describes a classification task in a closed-set scenario. Given a finite set of candidate authors and corresponding labeled texts, the objective is to determine which of the authors has written another set of anonymous or disputed texts.
In this work, we propose a probabilistic autoencoding framework to deal with this supervised classification task.
More precisely, we are extending a variational autoencoder (VAE) with embedded Gaussian mixture model to a Student-$t$ mixture model. 
Autoencoders have had tremendous success in learning latent representations. However, existing VAEs are currently still bound by limitations imposed by the assumed Gaussianity of the underlying probability distributions in the latent space. In this work, we are extending the Gaussian model for the VAE to a Student-$t$ model, which allows for an independent control of the ``heaviness'' of the respective tails of the implied probability densities. 
Experiments over an Amazon review dataset indicate superior performance of the proposed method.
\end{abstract}

\section{Introduction}
\input{introduction}

\section{Preliminaries}
\label{sec:pre}
\input{preliminaries}

\section{Model Description}
\label{sec:theory}
\input{theory}

\section{Evaluation}
\label{sec:results}
\input{results}

\section{Conclusion}
\label{sec:conclusion}
\input{conclusion}

\clearpage
\bibliographystyle{coling}
\bibliography{refs}

\end{document}

%% file: introduction.tex
\begin{wrapfigure}{r}{0.4\linewidth}
\centering
\includegraphics{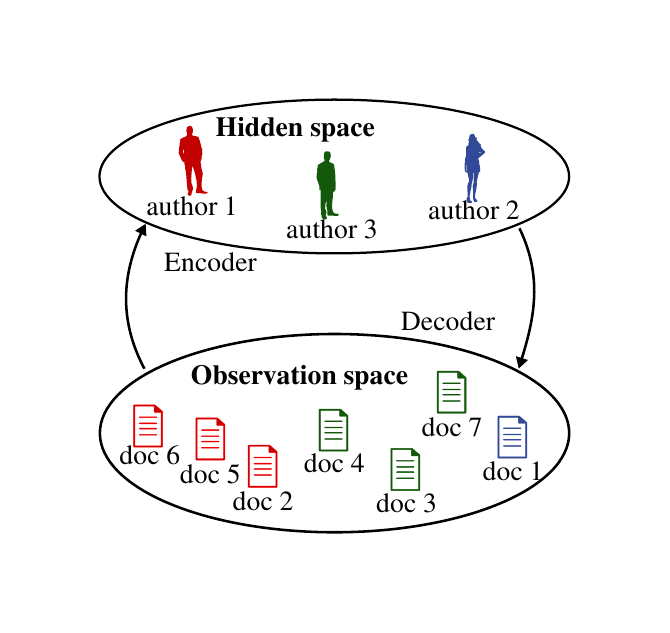}
\caption{Non-linear observation model for authorship analysis.}
\label{fig:latentspace}
\end{wrapfigure}

\textit{Supervised} authorship attribution traditionally refers to the task of analyzing the linguistic patterns of a text in order to determine who, from a finite set of enrolled authors, has written a document of unknown authorship. Nowadays, the focus of this \textit{closed-set} scenario has shifted from \textit{literary} to \textit{social media} authorship attribution, where methods have been developed to deal with large-scaled datasets of small-sized online texts~\cite{7555393}, \cite{ICASSP19}, \cite{8683747}, \cite{BigData19}\cite{tschuggnall-etal-2019-reduce}.

\textsc{AdHominem}, a linguistically motivated deep learning topology proposed in~\cite{BigData19}, can be seen as feature extractor for such tasks, where the resulting \textit{neural} feature vectors in the \textit{observation space} (see Fig.~\ref{fig:latentspace}) encode those stylistic characteristics of a text that are relevant for examining the writing style.
As described in~\cite{BigData19}, \textsc{AdHominem} was trained on a large Amazon review dataset considering $784.649$ different authors. 
Since short online texts (less than $1000$ tokens sample size) written by a huge amount of different authors are involved in fitting the \textsc{AdHominem} model, we may assume that these vectors do not (only) represent author-specific information but rather a general description of the writing style of a given text. 

In the case of a supervised authorship attribution task based on these features, it is beneficial to, first, map a neural feature vector onto a suitable representation in the \textit{latent space}, where it represents a distinguishable author-specific feature vector and then, second, to fit a (probabilistic) classifier that is embedded in this resulting latent space. 
We therefore present a new probabilistic autoencoding framework, incorporating a \gls{SMM} model into a \gls{VAE} framework to derive a joint learning mechanism for the latent manifold and its statistical representation.

The \gls{VAE} published by~\cite{KingmaVA} combines unsupervised deep learning with variational Bayesian methods. 
The \gls{VAE} framework relies on a probabilistic graphical model in the form of a directed acyclic graph, in which the hidden representations of an encoder network as well as the reconstructed outputs of a subsequent decoder network are treated as random variables.  
More precisely, the encoder defines a variational inference network, using high-dimensional observations to estimate an approximate posterior distribution in latent space,
and the decoder is a generative network, mapping latent representations back to distributions over the observation space.
The framework is used to generate compressed, approximate representations for virtually any type of patterned input. 
Depending on the targeted application, we may remove either the encoder or the decoder from the framework, once the joint training of the combined encoder-decoder system has been completed. 

The \gls{VAE} can be understood as a single-class probabilistic autoencoder since it is assumed that all latent representations are sampled from the same Gaussian distribution.
Different extensions of the conventional \gls{VAE} (e.g.~\cite{CVAE}), \cite{DilokthanakulMG16}), \cite{DeepLatent}, \cite{ladder}, \cite{svae}, \cite{StickBreaking}, \cite{Ebbers2017HiddenMM}, \cite{SIN}, \cite{S1}, \cite{s-vae18}, \cite{IW}, \cite{S2}) have been proposed. 
Particularly relevant to our work is the paper by~\cite{china}, in which the authors broadened the conventional VAE concept by generalizing the assumption of strictly Gaussian distributions to mixtures of Gaussians. This structure represents our baseline in the following.

In this paper, we are incorporating the assumption of Student-t distributed data into the joint learning mechanism for the latent manifold and its statistical representation. 
The advantage of using the Student-$t$ model is that we obtain a means to independently control the \textit{heaviness} of the respective tails of each distribution. Our generalization of the framework can be successfully employed in a variety of common machine learning tasks:
\begin{itemize}
 \item \textit{Unsupervised learning:} The basic architecture of our proposed method provides a generic recipe to autonomously group high-dimensional data into meaningful clusters.
 \item \textit{Supervised learning:} The derived loss function of our training method carries a cross-entropy term, which can be used to directly fuse class label information into the learning task. We are thereby able to enforce learning in a predefined/supervised direction as well.
 \item \textit{Semi-supervised learning:} In some cases we may have a large amount of training data, only a small subset of which is labeled. In this situation, we can utilize our method to, first, pre-train the model in a supervised manner and then refine the model with the unlabeled data in an unsupervised fashion. 
\end{itemize}

The remainder of the paper is divided into four sections. Section 2 discusses preliminary background information
and Section 3 describes the proposed method in detail. Experimental results are presented in Section 4 and a conclusion is offered in Section 5.

%% file: preliminaries.tex
\subsection{Variational Autoencoders}
On a very abstract level, a \gls{VAE} as described in~\cite{KingmaVA} is a neural-network-based technique for learning lower-dimensional latent representations given a set of higher-dimensional observable training samples.
Let $\mathcal{O} =\{ \boldsymbol{o}_n \}_{n=1}^N = \{\boldsymbol{o}_1, \ldots , \boldsymbol{o}_N\}$ denote a training set of high-dimensional observation vectors $\boldsymbol{o}_n\in \mathbb{R}^{L}$ for $n\in \{1\ldots N\}$. We assume that the $\boldsymbol{o}_n$ are independent and identically distributed samples from either a continuous or a discrete random variable. Furthermore, we use $\mathcal{X} = \{\boldsymbol{x}_1, \ldots , \boldsymbol{x}_N\}$ to denote a collection of $N$ low-dimensional latent representation vectors $\boldsymbol{x}_n\in \mathbb{R}^{D}$, where each $\boldsymbol{x}_n$ is associated with a corresponding observation $\boldsymbol{o}_n$. We are aiming to learn the marginal distribution of the observable variables
\begin{align}
    \label{eq:ml_of_ov}
    p_{\boldsymbol{\theta}}(\boldsymbol{o}_n) = \int p_{\boldsymbol{\theta}}(\boldsymbol{o}_n, \boldsymbol{x}_n)~ \mathrm{d}\boldsymbol{x}_n.
\end{align}
The joint distribution $p_{\boldsymbol{\theta}}(\boldsymbol{o}_n, \boldsymbol{x}_n)$ is parameterized by a framework-dependent coefficient/parameter vector $\boldsymbol{\theta}$.
The underlying \gls{VAE} framework that links the observations $\boldsymbol{o}_n$ with the latent vectors $\boldsymbol{x}_n$ consists of two neural networks, an encoder and a decoder.
Both, encoder and decoder networks are deterministic nonlinear functions whose outputs define the set of parameters 
which can be used to sample new observations or new latent representations. 
The joint distribution in Eq.~\ref{eq:ml_of_ov} can be factored as
$p_{\boldsymbol{\theta}}(\boldsymbol{o}_n, \boldsymbol{x}_n) = p_{\boldsymbol{\theta}}(\boldsymbol{o}_n|\boldsymbol{x}_n)~p(\boldsymbol{x}_n)$ which, in turn, justifies the following generative procedure: 
\begin{enumerate}[leftmargin=5mm, noitemsep]
 \item Sample a latent space representation $\boldsymbol{x}_n \sim p(\boldsymbol{x}_n)$,
	where
\begin{align}
	\label{eq:vae_enc_1}
    p(\boldsymbol{x}_n) = \mathcal{N}(\boldsymbol{x}_n |\boldsymbol{0}_{D \times 1}, \boldsymbol{I}_{D \times D})
\end{align}
and $\mathcal{N}(\cdot)$ defines the Gaussian distribution.
  Parameters $\boldsymbol{0}_{D \times 1}$ and $\boldsymbol{I}_{D \times D}$ indicate a $D$-dimensional zero mean vector and a unit covariance matrix, respectively.
\item Decode a parameter set for the $n$-th observation $\boldsymbol{o}_n$,
\begin{align}
      \label{eq:vae_enc_2}
      \{ \boldsymbol{\mu}_n^{(\boldsymbol{o}|\boldsymbol{x})}, \ln  \boldsymbol{\sigma}_n^{(\boldsymbol{o}|\boldsymbol{x})} \}  
		= \text{Decoder}_{\boldsymbol{\theta}} \big( \boldsymbol{x}_n \big).
\end{align}
All weights and bias terms of the neural network are contained in 
$\boldsymbol{\theta}$. 
\item Sample a new observation $\boldsymbol{o}_n \sim p_{\boldsymbol{\theta}}(\boldsymbol{o}_n|\boldsymbol{x}_n)$ with
\begin{align}
	\label{eq:vae_enc_3}
	  p_{\boldsymbol{\theta}}(\boldsymbol{o}_n|\boldsymbol{x}_n) 
	    = \mathcal{N}\big( \boldsymbol{o}_n \big| \boldsymbol{\mu}_n^{(\boldsymbol{o}|\boldsymbol{x})} , 
			      \text{diag}\big\{\big(\boldsymbol{\sigma}_n^{(\boldsymbol{o}|\boldsymbol{x})}\big)^2 \big\}\big).
\end{align}
\end{enumerate}
The generative model described by Eqs.~\ref{eq:vae_enc_1} to~\ref{eq:vae_enc_3} is employing a deep neural network mapping in Eq.~\ref{eq:vae_enc_2}. The posterior distribution is parametrized by 
the decoder neural network to learn a deterministic function that transforms the $n$-th latent variable $\boldsymbol{x}_n$ into the higher-dimensional observation space
of~$\boldsymbol{o}_n$. In order to learn the parameters $\boldsymbol{\theta}$ of the decoder, we are constructing a so-called {\em inference model\/} that is complementary to our generative model.
Because the marginal likelihood in Eq.~\ref{eq:ml_of_ov} is intractable (due to the non-linearity implied in Eq.~\ref{eq:vae_enc_2}), we are equally unable to solve for the posterior distribution of the latent variable 
$p_{\boldsymbol{\theta}}(\boldsymbol{x}_n|\boldsymbol{o}_n) = p_{\boldsymbol{\theta}}(\boldsymbol{o}_n, \boldsymbol{x}_n) / p_{\boldsymbol{\theta}}(\boldsymbol{o}_n)$. We may, however, approximate
the posterior distribution via $q_{\boldsymbol{\phi}}(\boldsymbol{x}_n|\boldsymbol{o}_n) \approx p_{\boldsymbol{\theta}}(\boldsymbol{x}_n|\boldsymbol{o}_n)$, in which $\boldsymbol{\phi}$ represents a set of \textit{inference} parameters. Similarly to the generative model, the inference process is characterized by a second neural network:
\begin{enumerate}[leftmargin=5mm, noitemsep]
 \item[4.] Decode a parameter set for the $n$-th latent variable,
\begin{align}
 \{ \boldsymbol{\mu}_n^{(\boldsymbol{x}|\boldsymbol{o})}, \ln  \boldsymbol{\sigma}_n^{(\boldsymbol{x}|\boldsymbol{o})} \} 
	    = \text{Encoder}_{\boldsymbol{\phi}} \big( \boldsymbol{o}_n \big).
\end{align}
\item[5.] Sample a latent variable $\boldsymbol{x}_n \sim q_{\boldsymbol{\phi}}(\boldsymbol{x}_n|\boldsymbol{o}_n)$, where
\begin{align}
\label{eq:posterior_VAE}
    q_{\boldsymbol{\phi}}(\boldsymbol{x}_n|\boldsymbol{o}_n) &= \mathcal{N}\big( \boldsymbol{x}_n \big| \boldsymbol{\mu}_n^{(\boldsymbol{x}|\boldsymbol{o})}, 
		    \text{diag}\big\{\big(\boldsymbol{\sigma}_n^{(\boldsymbol{x}|\boldsymbol{o})} \big)^2 \big\}\big).
\end{align}
\end{enumerate}

\subsection{Lower Bound and Re-parameterization Trick}
\label{sec:LBRT}
The variational Bayes approach is applied by simultaneously learning both the parameters of $p_{\boldsymbol{\theta}}(\boldsymbol{o}_n|\boldsymbol{x}_n)$ and those of the 
posterior approximation $q_{\boldsymbol{\phi}}(\boldsymbol{x}_n|\boldsymbol{o}_n)$. 
We can decompose the log-likelihood of the marginal distribution in Eq.~\ref{eq:ml_of_ov} to obtain an \gls{ELBO} with
$\mathcal{L}_{\boldsymbol{\theta}, \boldsymbol{\phi}} (\boldsymbol{o}_n) \le \ln p_{\boldsymbol{\theta}}(\boldsymbol{o}_n)$. Cumulatively, i.e. for all observations, we obtain 
\begin{align}
  \label{eq:vae_lowerbound}
 \mathcal{L}_{\boldsymbol{\theta}, \boldsymbol{\phi}}(\mathcal{O}) 
        =  \sum_{n=1}^N \mathcal{L}_{\boldsymbol{\theta}, \boldsymbol{\phi}}(\boldsymbol{o}_n).
\end{align}
The \gls{ELBO} for each individual observation is given by
\begin{align}
\mathcal{L}_{\boldsymbol{\theta}, \boldsymbol{\phi}} (\boldsymbol{o}_n)
\label{eq:vae_lb_interpret}
    &= \mathbb{E}_{q_{\boldsymbol{\phi}}(\boldsymbol{x}_n|\boldsymbol{o}_n)}
	  \bigg[  \ln \bigg\{\frac{p_{\boldsymbol{\theta}}(\boldsymbol{o}_n,\boldsymbol{x}_n)}{q_{\boldsymbol{\phi}}(\boldsymbol{x}_n|\boldsymbol{o}_n)} 
	  \bigg\} \bigg]
 \\ \nonumber
      &= \mathbb{E}_{q_{\boldsymbol{\phi}}(\boldsymbol{x}_n|\boldsymbol{o}_n)} \big[ \ln p_{\boldsymbol{\theta}}(\boldsymbol{o}_n | \boldsymbol{x}_n)\big]
	 -\text{KL} \big(q_{\boldsymbol{\phi}}(\boldsymbol{x}_n|\boldsymbol{o}_n) || p(\boldsymbol{x}_n) \big).
\end{align}
We use $\text{KL}(\cdot)$ to denote the \gls{KL} divergence.
The subscripts $\boldsymbol{\theta}$ and $\boldsymbol{\phi}$ of the lower bound in Eq.~\ref{eq:vae_lowerbound} denote the parameter space over which optimization is performed.
The first term in Eq.~\ref{eq:vae_lb_interpret} can be interpreted as a reconstruction measure describing the accuracy of the encoding-decoding framework. It can be approximated as
\begin{align}
\mathbb{E}_{q_{\boldsymbol{\phi}}(\boldsymbol{x}_n|\boldsymbol{o}_n)} \big[ \ln p_{\boldsymbol{\theta}}(\boldsymbol{o}_n | \boldsymbol{x}_n)\big]
\approx
\frac{1}{T} \sum_{t=1}^{T} \ln p_{\boldsymbol{\theta}}(\boldsymbol{o}_n|\boldsymbol{x}_{n,t}),
\nonumber
\end{align}
where $\boldsymbol{x}_{n,t} \sim q_{\boldsymbol{\phi}}(\boldsymbol{x}_n|\boldsymbol{o}_n)$.
The second term in Eq.~\ref{eq:vae_lb_interpret} can be seen as a regularizer, which attempts to maintain similarity between the approximated posterior and the prior.

Variational inference learning can now be accomplished through stochastic gradient descent.
Derivative calculations with respect to the parameters of a stochastic variable is accomplished through a commonly used re-parameterization trick. Consider the term $q_{\boldsymbol{\phi}}(\boldsymbol{x}_n|\boldsymbol{o}_n)$ as an example. In a first step we sample from a standard normal distribution, i.e. we consider~$\boldsymbol{\epsilon}_t \sim \mathcal{N}(\boldsymbol{0}_{D \times 1} \boldsymbol{I}_{D\times D})$. In a second step we transform the resulting random noise $\boldsymbol{\epsilon}_t$ 
via
\begin{align}
 \boldsymbol{\widetilde{x}}_{n,t} = \boldsymbol{\mu}_n^{(\boldsymbol{\boldsymbol{x}}|\boldsymbol{o})} 
		      + \boldsymbol{\sigma}_n^{(\boldsymbol{\boldsymbol{x}}|\boldsymbol{o})} \odot \boldsymbol{\epsilon}_t
\end{align}
to obtain a sample that will be distributed according to $q_{\boldsymbol{\phi}}(\boldsymbol{x}_n|\boldsymbol{o}_n)$, and which can nonetheless be differentiated with respect to the parameters $\boldsymbol{\mu}_n$ and $\boldsymbol{\sigma}_n$. Note that the symbol $\odot$ is used to indicate element-wise vector multiplication.

\subsection{Student-$t$ Distributions in Latent Space}
A Student-$t$ probability density function (PDF) is a unimodal PDF similar to a Gaussian, but with an additional parameter that controls the ``heaviness'' of its tails.
Following~\cite{Murphy}, we define the Student-$t$ distribution for the $n$-th latent representation
$\boldsymbol{x}_n$ by assuming that this $D$-dimensional vector belongs to the $k$-th cluster with $k \in \{1, \ldots K \}$ as
\begin{align}
    \mathcal{S}  (\boldsymbol{x}_n \big| \boldsymbol{\mu}_k, \boldsymbol{\Sigma}_k,\nu_k)
	= \frac{\Gamma(\tfrac{\nu_k+D}{2})}{\Gamma(\tfrac{\nu_k}{2})}  
	     \frac{\det(\boldsymbol{\Sigma}_k)^{-\tfrac{1}{2}}}{ (\pi \nu_k)^{\tfrac{D}{2}}} 
     \bigg[ 1 + \tfrac{1}{\nu_k} (\boldsymbol{x}_n - \boldsymbol{\mu}_k)^T \boldsymbol{\Sigma}_k^{-1} 
	    (\boldsymbol{x}_n - \boldsymbol{\mu}_k) \bigg]^{-\tfrac{\nu_k+D}{2}}
	    \nonumber
\end{align}
where $\boldsymbol{\mu}_k$ defines the $D$-dimensional mean vector of the $k$-th class, $\boldsymbol{\Sigma}_k$ denotes the 
$D\times D$ scale matrix and $\nu_k \in [0, \infty]$ is the number of degrees of freedom. 
For $\nu_k \rightarrow \infty$, the Student-$t$ distribution tends towards a Gaussian distribution of the same mean vector and covariance matrix.
Alternatively, we can understand the Student-$t$ distribution as the marginalization with respect to a hidden variable, i.e.
\begin{align}
  \label{eq:StudentOpen}
  \mathcal{S}&(\boldsymbol{x}_n \big| \boldsymbol{\mu}_k, \boldsymbol{\Sigma}_k, \nu_k)
    =  \int_{0}^{+\infty} \mathcal{N}\big(\boldsymbol{x}_n\big|\boldsymbol{\mu}_k, \boldsymbol{\Sigma}_k/u_{nk} \big) 
		~\mathcal{G}\big(u_{nk}  \big|\tfrac{\nu_k}{2}, \tfrac{\nu_k}{2} \big) ~\mathrm{d}u_{nk}
\end{align}
where $u_{nk} > 0$ is the hidden scale variable. The term $\mathcal{G}(\cdot)$ defines the Gamma distribution.
The normal distribution is defined as
\begin{align}
   \mathcal{N} \big(\boldsymbol{x}\big|\boldsymbol{\mu}, \boldsymbol{\Sigma}/u \big)
    =  (2 \pi)^{-\tfrac{D}{2}} ~\det( \boldsymbol{\Sigma}/u)^{-\tfrac{1}{2}} 
     \exp\bigg\{ -\tfrac{1}{2} (\boldsymbol{x} - \boldsymbol{\mu})^T \big(\boldsymbol{\Sigma}/u \big)^{-1} (\boldsymbol{x} - \boldsymbol{\mu}) \bigg\}
\end{align}
and the Gamma distribution is given by
\begin{align}
    \mathcal{G}\big(u \big|\alpha, \beta \big) = \frac{\beta^{\alpha}}{\Gamma(\alpha)}~ u^{\alpha -1} \exp\big\{ -\beta u \big\}
\end{align}
for $u>0$ and $\alpha,\beta > 0$.
\subsection{A Student-$t$ Mixture Model}
A finite \gls{SMM} is defined as a weighted sum of multivariate
Student-$t$ distributions. With $\,\sum_{k=1}^K \pi_k = 1\,$ we may write
\begin{align}
p(\boldsymbol{x_n}) &= \sum_{k=1}^K \Pr(z_{nk}=1) ~ p(\boldsymbol{x_n} |z_{nk}=1) 
= \sum_{k=1}^K \pi_k ~ \mathcal{S}(\boldsymbol{x}_n \big| \boldsymbol{\mu}_k, \boldsymbol{\Sigma}_k,\nu_k).
\label{eq:SMM} 
\end{align}
As mentioned by~\cite{Bishop_Paper} and~\cite{Archambeau}, we can view the Student-$t$ distribution in Eq.~\ref{eq:StudentOpen} 
as the marginalization of a Gaussian-Gamma distribution by integrating out the hidden scale variable $u_{nk}$.
This infinte mixture of normal distributions with the same mean 
vector but with different covariance matrices can be incorporated into a generative process.
Omitting the dependency on the hyper-parameters $\boldsymbol{\mu}_k, \boldsymbol{\Sigma}_k$ and $\nu_k$, Eq.~\ref{eq:StudentOpen} can be rewritten as
\begin{align}
\mathcal{S}  (\boldsymbol{x}_n \big| \boldsymbol{\mu}_k, \boldsymbol{\Sigma}_k, \nu_k)
   = p(\boldsymbol{x_n}|z_{nk}=1)
   = \int_0^{+\infty} p(u_{nk}|z_{nk}=1) ~p(\boldsymbol{x_n}| u_{nk},z_{nk}=1)~\mathrm{d}u_{nk}
\label{eq:inf_mix_w_uz}
\end{align}
where $z_{nk} \in \{0, 1\}$ is an indicator variable showing whether the $n$-th observation belongs to the $k$-th class. 
Consequently, our generative model in the latent space is augmented by the scale parameter $u_{nk}$ as an additional latent variable.

%% file: theory.tex
\subsection{The Generative Model}
\begin{figure}[t]
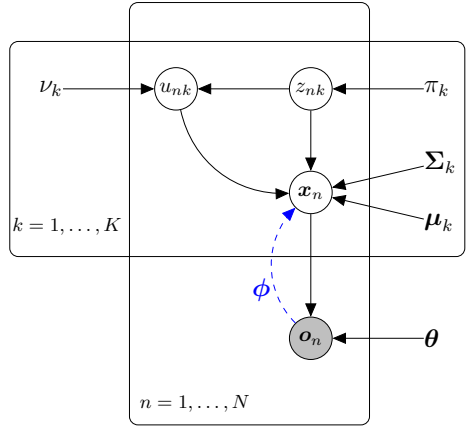

    \centering
    \scalebox{.8}{
    \tikz{
    \node[latent] (x) {$\boldsymbol{x}_n$} ;
    \node[latent, above=of x] (z) {$z_{nk}$} ;
    \node[latent, left=of z, xshift=-0.5cm] (u) {$u_{nk}$} ;
    \node[obs, below=of x, yshift=-0.7cm] (y) {$\boldsymbol{o}_n$} ; 
    \node[const, right=of y, xshift=0.5cm] (d) {\large$\boldsymbol{\theta}$} ; 
    \node[const, right=of z, xshift=0.5cm] (pi) {\large$\pi_k$} ; 
    \node[const, left=of u, xshift=-.5cm] (nu) {\large$\nu_k$} ; 
    \node[const, right=of x, xshift=0.5cm, yshift=0.5cm] (sigma) {\large$\boldsymbol{\Sigma}_k$} ; 
    \node[const, right=of x, xshift=0.5cm, yshift=-0.5cm] (mu) {\large$\boldsymbol{\mu}_k$} ; 
    \node[const, left=of y, xshift=.7cm, yshift=.8cm, blue] (enc) {\large$\boldsymbol{\phi}$} ; 
    \plate[minimum width =2cm, minimum height=7cm, inner sep=.5cm, xshift=.1cm, yshift=0.11cm] {plate1} {(u) (x) (y) (z)} {}; 
    \plate[minimum width =7cm, inner sep=0.3cm, xshift=-0.17cm, yshift=0.12cm] {plate2} {(u) (z) (pi) (mu) (sigma) (nu)} {}; 
    \node [below,xshift = 0.1cm, yshift = -2.87cm, xshift=-1cm] at (plate1) {\small $n=1,\ldots, N$};
    \node [below,xshift = 0.2cm, yshift = -1cm, xshift=-3cm] at (plate2) {\small $k=1, \ldots, K$};
    \edge {pi} {z} ;
    \edge {z, nu} {u} ;
    \edge {z, sigma} {x} ;
    \edge {x, d} {y} ;
    \edge {mu} {x} ;
    \draw[->] (u) edge [bend right=40] (x) ; 
    \draw[->, dashed, blue] (y) edge [bend left=45] (x) ; 
    }}
    \caption{Probabilistic graphical model for the generative process of the proposed Student-$t$ VAE (tVAE).}
    \label{fig:pgm}
\end{figure}

We use $\boldsymbol{\xi} 
	= \{\boldsymbol{\mu}, \boldsymbol{\nu}, \boldsymbol{\Sigma}, \boldsymbol{\pi} \} = \{\boldsymbol{\mu}_k, \nu_k, \boldsymbol{\Sigma}_k,\pi_k \}_{k=1}^K
$ to denote the set of all hyper-parameters of an SMM after Eq.~\ref{eq:SMM}. In a process similar to the one outlined in 
Section~\ref{sec:LBRT} for a VAE, we can generate observation samples $\boldsymbol{o}_n$ for the proposed tVAE with the following $5$ steps:
\begin{enumerate}[leftmargin=5mm, noitemsep]
  \item Choose a cluster for the $n$-th observation by sampling the one-hot vector $\boldsymbol{z}_n \sim p(\boldsymbol{z}_n)$, where
\begin{align}
	    p_{\boldsymbol{\xi}}(\boldsymbol{z}_n) = \prod_{k=1}^K \pi_k^{z_{nk}}
\end{align}
and $\boldsymbol{z}_n = \big[ z_{n1}, \ldots , z_{nK} \big]^T$.
\item Sample the $n$-th scale vector $\boldsymbol{u}_n \sim p_{\boldsymbol{\xi}}(\boldsymbol{u}_n|\boldsymbol{z}_n)$, where
\begin{align}
	    \label{eq:gp_u}
	    p_{\boldsymbol{\xi}}(\boldsymbol{u}_n|\boldsymbol{z}_n) = \prod_{k=1}^K \mathcal{G}\big(u_{nk} |\tfrac{\nu_k}{2}, \tfrac{\nu_k}{2} \big)^{z_{nk}}
	\end{align}
and $\boldsymbol{u}_n = \big[ u_{n1}, \ldots , u_{nK} \big]^T$.
\item Sample a new latent representation for the $n$-th observation, $\boldsymbol{x}_n \sim p_{\boldsymbol{\xi}}(\boldsymbol{x}_n|\boldsymbol{z}_n, \boldsymbol{u}_n)$ with 
 \begin{align}
	\label{eq:gp_x}
  	p_{\boldsymbol{\xi}}(\boldsymbol{x}_n|\boldsymbol{z}_n, \boldsymbol{u}_n) = \prod_{k=1}^K  \mathcal{N}(\boldsymbol{x}_n | \boldsymbol{\mu}_k, \boldsymbol{\Sigma}_k / u_{nk}  )^{z_{nk}}.
 \end{align}
\item Decode a parameter set for the $n$-th observation $\boldsymbol{o}_n$,
       \begin{align}
	   \big\{\boldsymbol{\mu}_{n}^{(\boldsymbol{o}|\boldsymbol{x})}, \log \boldsymbol{\sigma}_{n}^{(\boldsymbol{o}|\boldsymbol{x})}
	   \big\} 
		= \text{Decoder}_{\boldsymbol{\theta}}(\boldsymbol{x}_n).
 \end{align}
 The set $\boldsymbol{\theta}$ summarizes all weights and bias terms of the decoder network.
\item Sample an observation $\boldsymbol{o}_n \sim p_{\boldsymbol{\theta}}(\boldsymbol{o}_n|\boldsymbol{x}_n)$, where   
\begin{align}
	p_{\boldsymbol{\theta}}(\boldsymbol{o}_n|\boldsymbol{x}_n) 
	  = \mathcal{N}\big(\boldsymbol{o}_n|\boldsymbol{\mu}_{n}^{(\boldsymbol{o}|\boldsymbol{x})}, 
                \boldsymbol{\Sigma}_{n}^{(\boldsymbol{o}|\boldsymbol{x})}
                \big)
\end{align}
with $\,\,\boldsymbol{\Sigma}_{n}^{(\boldsymbol{o}|\boldsymbol{x})} = \text{diag} \big\{ \big( \boldsymbol{\sigma}_{n}^{(\boldsymbol{o}|\boldsymbol{x})} \big)^2 \big\}$.
\end{enumerate}
An illustration of the generative process for the proposed Student-$t$ VAE with a corresponding graphical model is shown in 
Fig.~\ref{fig:pgm}.

\subsection{Approximate Inference}
At this point it is notationally beneficial to define the set
$\mathcal{H} 
	  = \big\{ \mathcal{Z}, \mathcal{U}, \mathcal{X} \big\} 
	  = \big\{ \{z_{nk}, u_{nk}\}_{k=1}^K, \boldsymbol{x}_n  \big\}_{n=1}^N
$ of all latent variables of our proposed framework. We apply the mean-field approximation
to find an analytical expression of the approximate joint posterior distribution $q_{\boldsymbol{\phi}}(\mathcal{H})$. The symbol $\boldsymbol{\phi}$ is used to represent the set of all weights and bias terms of the underlying encoder network. 
Suppose, the joint posterior distribution of $\mathcal{H}$ can be factored such that
\begin{align}
  \label{eq:fac_H}
    q_{\boldsymbol{\phi}}(\mathcal{H}|\mathcal{O}) = \prod_{ i} q_{\boldsymbol{\phi}}(\mathcal{H}_i|\mathcal{O}),
\end{align}
then the  posterior distribution can be obtained after~\cite{Bishop_Book}~from:
\begin{align}
    \label{eq:UpdRule}
   \ln q_{\boldsymbol{\phi}}(\mathcal{H}_j|\mathcal{O}) 
	  = \mathbb{E}_{\prod_{i\ne j} q_{\boldsymbol{\phi}}(\mathcal{H}_i|\mathcal{O})} \big[ \ln p(\mathcal{O}, \mathcal{H})\big]+\text{const}.
\end{align}
In our context, the product $\prod_{ i} q(\mathcal{H}_i|\mathcal{O})$ in Eq.~\ref{eq:fac_H} represents a suitable factorization of the joint posterior distribution of all latent variables.
One possible approximate factorization is:
\begin{align}
        \label{eq:mean_field}
    q_{\boldsymbol{\phi}}(\mathcal{H}|\mathcal{O}) 
   \approx \prod_{n=1}^N q_{\boldsymbol{\phi}}(\boldsymbol{x}_n|\boldsymbol{o}_n) \prod_{k=1}^K q( z_{nk}, u_{nk}). \quad
\end{align}
The employed generative model implies that there is a statistical dependency between $\boldsymbol{x}_n$ and $\boldsymbol{z}_n, \boldsymbol{u}_n$. It can be argued, however, that we may ignore this dependency in our case because the posterior distribution in the latent space is encoded by the second neural network, i.e.
\begin{align}
    q_{\boldsymbol{\phi}}(\boldsymbol{x}_n|\boldsymbol{o}_n) 
    &= \mathcal{N}(\boldsymbol{x}_n | \boldsymbol{\mu}^{(\boldsymbol{x}|\boldsymbol{o})}_n, \boldsymbol{\Sigma}^{(\boldsymbol{x}|\boldsymbol{o})}_n ), 
    \\ \text{with}\,\,\,\,\, 
    \{ \boldsymbol{\mu}_n^{(\boldsymbol{x}|\boldsymbol{o})}, & \log \boldsymbol{\sigma}^{(\boldsymbol{x}|\boldsymbol{o})}_n \} 
    = \text{Encoder}_{\boldsymbol{\phi}}(\boldsymbol{o}_n ),
\end{align}
where, again, $\boldsymbol{\Sigma}^{(\boldsymbol{x}|\boldsymbol{o})}_n = \text{diag}\big\{ \big(\boldsymbol{\sigma}^{(\boldsymbol{x}|\boldsymbol{o})}_n \big)^2 \big\}$.
Note that the posterior distribution of $\boldsymbol{z}_n$ and $\boldsymbol{u}_n$ in Eq.~\ref{eq:mean_field} does not directly depend on $\boldsymbol{\phi}$, which is important for the calculation of the loss function discussed in Section~\ref{sec:loss_func}.
It is not necessary to approximate the joint posterior distribution of $\boldsymbol{z}_n$ and $\boldsymbol{u}_n$ because
it is possible to analytically determine the marginal distributions $q(z_{nk})$ and $q(u_{nk}|z_{nk})$ given the joint distribution
$q(z_{nk}, u_{nk})$. 
Applying Eq.~\ref{eq:UpdRule}, the optimal joint distribution $q(z_{nk}, u_{nk})$ yields:
\begin{align}
 \label{eq:q_zu}
    \ln q(z_{nk}, u_{nk}) 
    = z_{nk} \bigg( \ln \pi_k + \ln \mathcal{G}\big(u_{nk} |\tfrac{\nu_k}{2}, \tfrac{\nu_k}{2} \big)
        +\mathbb{E}_{q_{\boldsymbol{\phi}}(\boldsymbol{x}_n|\boldsymbol{o}_n)} 
			\big[  \ln \mathcal{N}(\boldsymbol{x}_n | \boldsymbol{\mu}_k,  \boldsymbol{\Sigma}_k /u_{nk} ) \big] \bigg) 
          +\text{const}.
\end{align}
For the marginal distribution $q(z_{nk}=1)$ we have
\begin{align}
\label{eq:mag_dib_1}
    q(z_{nk}=1) 
 &= \int_0^{+\infty} q(z_{nk}=1, u_{nk}) ~\mathrm{d}u_{nk}
 	 \nonumber \\    & \propto \int_0^{+\infty} \pi_k ~ \mathcal{G}\big(u_{nk} |\tfrac{\nu_k}{2}, \tfrac{\nu_k}{2} \big)
 \times \exp \big\{\mathbb{E}_{q(\boldsymbol{x}_n)} \big[  \ln \mathcal{N}(\boldsymbol{x}_n | \boldsymbol{\mu}_k,   \boldsymbol{\Sigma}_k /u_{nk}) \big]
	 \big\}~\mathrm{d}u_{nk}.
\end{align}
In a first step, we may recast Eq.~\ref{eq:mag_dib_1} into a form given by Eq.~\ref{eq:post_z_int}.
 \begin{align}
  \label{eq:post_z_int}
q(z_{nk}=1) &\propto \pi_k ~
      \bigg(\frac{ \nu_k }{ \nu_k + \mathrm{Tr}\big\{ \boldsymbol{\Sigma}^{(\boldsymbol{x}|\boldsymbol{o})}_n  \boldsymbol{\Sigma}_k^{-1}\big\}  } \bigg)^{\tfrac{\nu_k}{2}}
 	 \nonumber \\ & \quad\times
  \int_{0}^{+\infty} 
	  \mathcal{G}\big(u_{nk} \big|\tfrac{\nu_k}{2} , \frac{\nu_k +  \mathrm{Tr}\big\{ \boldsymbol{\Sigma}^{(\boldsymbol{x}|\boldsymbol{o})}_n  \boldsymbol{\Sigma}_k^{-1} \big\}}{2}\big)
	  ~\mathcal{N}\big(\boldsymbol{\mu}_n^{(\boldsymbol{x}|\boldsymbol{o})} \big| \boldsymbol{\mu}_k, \boldsymbol{\Sigma}_k / u_{nk} \big) 
	  ~ \mathrm{d}u_{nk}
\end{align}
Obviously, the expression of the posterior probabilities in Eq.~\ref{eq:post_z_int} can not be implemented directly, but it provides an approach to interpret the posterior probabilities. For instance, the integral represents an infinite mixture model which can be seen as a
probability model for the estimated mean vector $\boldsymbol{\mu}_n^{(\boldsymbol{x}|\boldsymbol{o})}$. Note that the mean vector is part of the output of the decoder. 
Instead of a simple point estimator for the scale parameter $u_{nk}$, the mixture model utilizes all possible values. Besides the mean vector, the expression also considers the estimated 
covariance matrix  
$\boldsymbol{\Sigma}_n^{(\boldsymbol{x}|\boldsymbol{o})}$. The terms in front of the integral are class-specific weights, where the trace of the matrix multiplication,  
$\mathrm{Tr}\big( \boldsymbol{\Sigma}^{(\boldsymbol{x}|\boldsymbol{o})}_n  \boldsymbol{\Sigma}_k^{-1} \big)$, is always $\ge 0$ and can be rewritten as the 
expectation of the Mahalanobis distance,
\begin{align}
\mathrm{Tr}&\big\{ \boldsymbol{\Sigma}^{(\boldsymbol{x}|\boldsymbol{o})}_n  \boldsymbol{\Sigma}_k^{-1} \big\}
 = \mathbb{E}_{q_{\boldsymbol{\phi}}(\boldsymbol{x}_n|\boldsymbol{o}_n) }\big[  (\boldsymbol{x}_n - \boldsymbol{\mu}^{(\boldsymbol{x}|\boldsymbol{o})}_n)^T
	    \boldsymbol{\Sigma}_k^{-1} (\boldsymbol{x}_n - \boldsymbol{\mu}^{(\boldsymbol{x}|\boldsymbol{o})}_n) \big].
\nonumber
\end{align}
The next step now is to find a closed form expression for the optimal posterior distribution $q(z_{nk}=1)$. We can rearrange Eq.~\ref{eq:post_z_int} such that
\begin{align}
  q(z_{nk}=1)
  &\propto
   \pi_k~  \frac{(\tfrac{\nu_k}{2})^{\tfrac{\nu_k}{2}}}{\Gamma(\tfrac{\nu_k}{2})} ~ \det(\boldsymbol{\Sigma}_k )^{-\tfrac{1}{2}}
\nonumber\\ &\quad    
	      \times	  \int_{0}^{+\infty} (u_{nk})^{\tfrac{\nu_k+D}{2}-1}
   \exp \big\{ u_{nk} \big[ \tfrac{\nu_k}{2}  +\tfrac{1}{2}  \mathrm{Tr}\big\{
	     \boldsymbol{\Sigma}^{(\boldsymbol{x}|\boldsymbol{o})}_n  \boldsymbol{\Sigma}_k^{-1} \big\}
  \nonumber \\ 
  \label{eq:qzu_rewritten}
  &  \quad	  
	    +\tfrac{1}{2} (\boldsymbol{\mu}_n^{(\boldsymbol{x}|\boldsymbol{o})} - \boldsymbol{\mu}_k)^T \boldsymbol{\Sigma}_k^{-1} 
	    (\boldsymbol{\mu}_n^{(\boldsymbol{x}|\boldsymbol{o})} - \boldsymbol{\mu}_k) \big] 
	    ~ \big\}  ~\mathrm{d} u_{nk}.
\end{align}
The integral in Eq.~\ref{eq:qzu_rewritten} has the following form:
\begin{align}
    \int_{0}^{+\infty} 
	x^{\alpha_k-1} e^{-u_{nk} ~\beta_{nk} } ~\mathrm{d}u_{nk},
\end{align}
where
\begin{align}
    \alpha_k &= \tfrac{\nu_k+D}{2}  \quad\text{and}
    \\ \beta_{nk} &= \tfrac{1}{2} \big[ (\boldsymbol{\mu}^{(\boldsymbol{x}|\boldsymbol{o})}_n - \boldsymbol{\mu}_k)^T \boldsymbol{\Sigma}_k^{-1} 
			  (\boldsymbol{\mu}^{(\boldsymbol{x}|\boldsymbol{o})}_n - \boldsymbol{\mu}_k) 
	+\nu_k  +  \mathrm{Tr}\big\{ \boldsymbol{\Sigma}^{(\boldsymbol{x}|\boldsymbol{o})}_n  \boldsymbol{\Sigma}_k^{-1} \big\} \big].
\end{align}
Substitution with $z = \beta_{nk} ~x$ results in
\begin{align}
    \beta_{nk}^{-\alpha_k} \int_{0}^{+\infty}  z^{\alpha_k-1} e^{-z} ~\mathrm{d}z =  \beta_{nk}^{-\alpha_k} ~\Gamma(\alpha_k).
\end{align}
This expression corresponds to the definition of the Gamma function $\Gamma(\cdot)$ for positive real-valued numbers~\cite{Bishop_Book}, which is true for $\alpha_k= \tfrac{\nu_k+D}{2} $.
Thus, the posterior distribution can be simplified as
\begin{align}
\label{eq:q_z}
    q(z_{nk}=1)
    &\propto  \pi_k~  
	\frac{\big(\tfrac{\nu_k}{2}\big)^{\tfrac{\nu_k}{2}}}{\Gamma(\tfrac{\nu_k}{2}) }  
	~\det(\boldsymbol{\Sigma}_k)^{-\tfrac{1}{2}}
    \Gamma\big(\alpha_k \big)  \beta_{nk}^{~-\alpha_k}.
\end{align}
Note that the last expression is similar to the closed form definition of the Student-$t$ distribution.
We can now determine the posterior distribution $q(u_{nk}|z_{nk}=1)$ using Eqs.~\ref{eq:q_zu} and~\ref{eq:q_z},
\begin{align}
 q(u_{nk}|z_{nk}=1) 
 = \frac{q(u_{nk}, z_{nk}=1)}{q(z_{nk}=1)}
 = \mathcal{G}\big(u_{nk} \big| \alpha_k, \beta_{nk} \big).
\end{align}
Consequently, $\alpha_k$ and $\beta_{nk}$ define the set of hyper-parameters for $ q(u_{nk}|z_{nk}=1)$.

\subsection{The Variational Lower Bound and the Loss Function}
\label{sec:loss_func}

We obtain a loss function for our proposed parameter optimization through the calculation of a variational lower bound. Given the observations, the optimal posterior probabilities, and the hyper-parameters, the lower bound for our probabilistic framework is given by
\begin{align}
\mathcal{L}_{\boldsymbol{\phi},\boldsymbol{\theta},\boldsymbol{\xi}}(\mathcal{O}) 
	  &= \mathbb{E}_{q_{\boldsymbol{\phi}}(\mathcal{X}, \mathcal{Z}, \mathcal{U}|\mathcal{O}) } 
	      \big[ \ln \big\{ \frac{p_{\boldsymbol{\theta,}\boldsymbol{\xi}}(\mathcal{Z}, \mathcal{U}, \mathcal{X}, \mathcal{O})}
		{q_{\boldsymbol{\phi}}(\mathcal{X}, \mathcal{Z}, \mathcal{U}|\mathcal{O})} \big\}  \big]
	\nonumber  \\ & = \sum_{n=1}^N  \mathbb{E}_{q_{\boldsymbol{\phi}}(\boldsymbol{x}_n,\boldsymbol{z}_n, \boldsymbol{u}_n|\boldsymbol{o}_n) } 
	      \big[ \ln \big\{ \frac{p_{\boldsymbol{\theta},\boldsymbol{\xi}}(\boldsymbol{x}_n, \boldsymbol{z}_n, \boldsymbol{u}_n, \boldsymbol{o}_n)}
		{q_{\boldsymbol{\phi}}(\boldsymbol{x}_n, \boldsymbol{z}_n, \boldsymbol{u}_n|\boldsymbol{o}_n)} \big\}  \big]
	\nonumber  \\& =  \sum_{n=1}^N \mathcal{L}_{\boldsymbol{\phi},\boldsymbol{\theta},\boldsymbol{\xi}}(\boldsymbol{o}_n).
\end{align}
The lower bound for the $n$-th observation can, thus, be broken down into the following 6 parts:
\begin{align}
  \mathcal{L}_{\boldsymbol{\phi},\boldsymbol{\theta},\boldsymbol{\xi}}(\boldsymbol{o}_n)
        &= \label{eq:lb_pz} \mathbb{E}_{q(\boldsymbol{z}_n)} \big[ \ln p_{\boldsymbol{\xi}}(\boldsymbol{z}_n) \big]
     \\& \quad \label{eq:lb_puz} + \mathbb{E}_{q(\boldsymbol{u}_n, \boldsymbol{z}_n)} \big[ \ln p_{\boldsymbol{\xi}}(\boldsymbol{u}_n|\boldsymbol{z}_n)  \big]
     \\& \quad \label{eq:lb_ppp} + \mathbb{E}_{q_{\boldsymbol{\phi}}(\boldsymbol{x}_n|\boldsymbol{o}_n) q(\boldsymbol{u}_n, \boldsymbol{z}_n)} 
		\big[ \ln p_{\boldsymbol{\xi}}(\boldsymbol{x}_n| \boldsymbol{z}_n, \boldsymbol{u}_n) \big]
     \\&  \label{eq:lb_obs} 
	  \quad + \mathbb{E}_{q_{\boldsymbol{\phi}}(\boldsymbol{x}_n|\boldsymbol{o}_n)} 
		\big[ \ln p_{\boldsymbol{\theta}}(\boldsymbol{o}_n|\boldsymbol{x}_n) \big]
     \\&  \label{eq:lb_qx} 
	\quad - \mathbb{E}_{q_{\boldsymbol{\phi}}(\boldsymbol{x}_n|\boldsymbol{o}_n)} \big[ \ln q_{\boldsymbol{\phi}}(\boldsymbol{x}_n|\boldsymbol{o}_n) \big]
     \\ \label{eq:lb_const} 
     &  \quad - \mathbb{E}_{q(\boldsymbol{u}_n, \boldsymbol{z}_n)} \big[ \ln q(\boldsymbol{u}_n, \boldsymbol{z}_n) \big].
\end{align}
For the sake of clarity we will discuss each term of the above expression separately. First, we may note that Term~\ref{eq:lb_const} remains constant during the gradient-based update phase, i.e. we have
\begin{align}
   \mathbb{E}_{q(\boldsymbol{u}_n, \boldsymbol{z}_n)} \big[ \ln q(\boldsymbol{u}_n, \boldsymbol{z}_n) \big] = \text{const},
\end{align}
since there is no dependency on the update parameters in $\boldsymbol{\theta}$, $\boldsymbol{\phi}$ and $\boldsymbol{\xi}$.
By assuming ergodicy, we can make the following approximation for Term~\ref{eq:lb_obs}:
\begin{align}
    \mathbb{E}_{q_{\boldsymbol{\phi}}(\boldsymbol{x}_n|\boldsymbol{o}_n)}  \big[ \ln p(\boldsymbol{o}_n|\boldsymbol{x}_n) \big] 
	 &\approx \frac{1}{T} \sum_{t=1}^T \ln p(\boldsymbol{o}_n|\boldsymbol{x}_{n,t}),\quad \boldsymbol{x}_{n,t} \sim q_{\boldsymbol{\phi}}(\boldsymbol{x}_n|\boldsymbol{o}_n) 
	 \nonumber \\   &\approx \frac{1}{T} \sum_{t=1}^T \ln 
			\mathcal{N}(\boldsymbol{o}_n |\boldsymbol{\mu}_{n,t}^{(\boldsymbol{o}|\boldsymbol{x})}, \boldsymbol{\Sigma}_{n,t}^{(\boldsymbol{o}|\boldsymbol{x})}
			).
\end{align}
For the re-parametrization trick, $\boldsymbol{x}_{n,t}$ is obtained as follows:
\begin{align}
  \boldsymbol{\epsilon}_t &\sim \mathcal{N}(\boldsymbol{0}_{D \times 1},\boldsymbol{I}_{D \times D}) ,
 \\ \boldsymbol{x}_{n,t} 
	      &= \boldsymbol{\mu}_n^{(\boldsymbol{\boldsymbol{x}}|\boldsymbol{o})} 
		      + \boldsymbol{\sigma}_n^{(\boldsymbol{\boldsymbol{x}}|\boldsymbol{o})} \odot \boldsymbol{\epsilon}_t,
\end{align}
which is fed into the decoder,
\begin{align}
 \{ \boldsymbol{\mu}_{n,t}^{(\boldsymbol{o}|\boldsymbol{x})}, \log  \boldsymbol{\sigma}_{n,t}^{(\boldsymbol{o}|\boldsymbol{x})},
 \nu_{n}^{(\boldsymbol{o}|\boldsymbol{x})}\}  
		&= \text{Decoder}_{\boldsymbol{\theta}} \big( \boldsymbol{x}_{n,t}\big).
\end{align}
Considering Term~\ref{eq:lb_qx} we may exploit the entropy of multivariate Gaussian distributions, i.e.
\begin{align}
-\mathbb{E}_{q_{\boldsymbol{\phi}}(\boldsymbol{x}_n|\boldsymbol{o}_n)} &\big[ \ln q_{\boldsymbol{\phi}}(\boldsymbol{x}_n|\boldsymbol{o}_n) \big]
	 = H\big( \mathcal{N}(\boldsymbol{x}_n |\boldsymbol{\mu}_n^{(\boldsymbol{x} |\boldsymbol{o})}, 
	\boldsymbol{\Sigma}_n^{(\boldsymbol{x}|\boldsymbol{o})}, \nu_n^{(\boldsymbol{x}|\boldsymbol{o})})\big).
\end{align}
Term~\ref{eq:lb_pz} can be summarized as
\begin{align}
   \mathbb{E}_{q(\boldsymbol{z}_n)} \big[ \ln p_{\boldsymbol{\xi}}(\boldsymbol{z}_n) \big]
= \sum_{k=1}^K  \underbrace{\mathbb{E}_{q(z_{nk})} \big[  z_{nk}\big]}_{=~\gamma_{nk}} \ln  \pi_k,
\end{align}
in which $\gamma_{nk}$ describes the posterior class probabilities such that
\begin{align}
    q(\boldsymbol{z}_n) = \prod_{k=1}^K \gamma_{nk}^{z_{nk}}
~ \text{ with } ~
    \gamma_{nk} = \frac{q(z_{nk}=1)}{\sum_{ k'} q(z_{nk'}=1)}.
    \nonumber
\end{align}
We may use Eq.~\ref{eq:q_z}
to compute $\gamma_{nk}$ since Eq.~\ref{eq:q_z}
only represents the unnormalized posterior probabilities.
For Term~\ref{eq:lb_puz}, it follows that 
\begin{align}
 \mathbb{E}_{q(\boldsymbol{u}_n, \boldsymbol{z}_n)} \big[ \ln p_{\boldsymbol{\xi}}(\boldsymbol{u}_n|\boldsymbol{z}_n)  \big] 
 &= \sum_{k=1}^K \gamma_{nk}~ \big[\mathbb{E}_{q(u_{nk}| z_{nk}=1)} \big[
	  \ln  \mathcal{G}\big(u_{nk} |\tfrac{\nu_k}{2}, \tfrac{\nu_k}{2} \big)  \big]\big]
 \nonumber  \\&= \sum_{\forall k}
      \gamma_{nk}~ 
      \big(
	\ln (\frac{\nu_k}{2})^{\frac{\nu_k}{2}} - \ln \Gamma(\frac{\nu_k}{2})
 \nonumber \\ &    \qquad\quad
	  +\big(\tfrac{\nu_k}{2}-1\big) \underbrace{\mathbb{E}_{q(u_{nk}|z_{nk=1})} \big[ \ln u_{nk} \big]}
	    _{=~\psi(\alpha_k)- \ln \beta_{nk}}
  -\tfrac{\nu_k}{2}  \underbrace{\mathbb{E}_{q(u_{nk}|z_{nk=1})} \big[ u_{nk} \big] }
	      _{=~ \tfrac{\alpha_k}{\beta_{nk}}}
	       \big)
\end{align}
where $\psi(\cdot)$ denotes the Digamma function~\cite{Bishop_Book}. To conclude, term~\ref{eq:lb_ppp} can be decomposed as follows:
\begin{align}
\mathbb{E}_{q_{\boldsymbol{\phi}}(\boldsymbol{x}_n|\boldsymbol{o}_n) q(\boldsymbol{u}_n, \boldsymbol{z}_n)} \big[ \ln p_{\boldsymbol{\xi}}(\boldsymbol{x}_n| \boldsymbol{z}_n, \boldsymbol{u}_n) \big]
   &=  
    \sum_{k=1}^K \gamma_{nk}~\bigg(
	    -\tfrac{D}{2} \ln \big\{ 2 \pi \big\}
	    - \tfrac{1}{2} \ln \det (\boldsymbol{\Sigma}_k)  
  + \tfrac{D}{2}  \big[\psi(\alpha_k) - \ln \beta_{nk} \big] 
  \nonumber \\ & \qquad\quad   - \tfrac{\alpha_k}{2~\beta{nk}} \big(
		    \text{Tr} \big\{\boldsymbol{\Sigma}_n^{(\boldsymbol{x}|\boldsymbol{o})} \boldsymbol{\Sigma}_k ^{-1} \big\}
 + (\boldsymbol{\mu}_n^{(\boldsymbol{x}|\boldsymbol{o})} - \boldsymbol{\mu}_k)^T
		  \boldsymbol{\Sigma}_k ^{-1}
		  (\boldsymbol{\mu}_n^{(\boldsymbol{x}|\boldsymbol{o})} - \boldsymbol{\mu}_k)
		  \big)
	  \bigg)
	  \nonumber.
\end{align}
Following~\cite{KingmaVA}, the negative of the derived evidence lower bound provides a loss function, i.e.
\begin{align}
J_{\boldsymbol{\phi},\boldsymbol{\theta},\boldsymbol{\xi}}\big(\mathcal{O} \big)
 = -\frac{1}{N} \sum_{n=1}^{N}  \mathcal{L}_{\boldsymbol{\phi},\boldsymbol{\theta},\boldsymbol{\xi}}(\boldsymbol{o}_n).
\end{align}

\subsection{Interpretation of the Lower Bound}
As a result, the ELBO for the $n$-th observation defined through Terms~\ref{eq:lb_pz} to~\ref{eq:lb_const} can be rewritten. $\,\,$We obtain the following, more compact expression:
\begin{align}
\label{eq:lb_underst}
\mathcal{L}_{\boldsymbol{\phi},\boldsymbol{\theta},\boldsymbol{\xi}}\big(\boldsymbol{o}_n \big)
&=   \frac{1}{T} \sum_{t=1}^T \ln 
			\mathcal{N}(\boldsymbol{o}_n |\boldsymbol{\mu}_{n,t}^{(\boldsymbol{o}|\boldsymbol{x})}, \boldsymbol{\Sigma}_{n,t}^{(\boldsymbol{o}|\boldsymbol{x})} )
\nonumber \\ & \quad + H\big(\mathcal{N}(\boldsymbol{x}_n |\boldsymbol{\mu}_n^{(\boldsymbol{x}|\boldsymbol{o})}, \boldsymbol{\Sigma}_n^{(\boldsymbol{x}|\boldsymbol{o})} )\big) 
\nonumber \\ & \quad + \sum_{k=1}^K \gamma_{nk} \ln \rho_{nk},
\end{align}
where
\begin{align}
\ln \rho_{nk} = 
       \ln q(z_{nk}=1) -H\big(\mathcal{G}(u_{nk}|\alpha_k, \beta_{nk})\big)  
	  -\tfrac{D}{2} \ln \big\{ 2 \pi \big\}.
\end{align}
The term $H\big(\mathcal{G}(u_{nk}|\alpha_k, \beta_{nk})\big)$ represents the entropy of the Gamma distribution with parameters $\alpha_k$ and~$\beta_{nk}$.
Similarly to the ELBO of the conventional VAE in Eq.~\ref{eq:vae_lb_interpret}, we may interpret the function of each term of the derived lower bound 
in Eq~\ref{eq:lb_underst}. The first term represents the reconstruction error, which measures how well the encoder-decoder framework fits the dataset. 
The second term can be seen as a regularizer quantifying the output of the decoder. Following the maximum entropy principle,
it will maximize the uncertainty with regard to possibly missing information.
The third term, the cross-entropy, evaluates the clustering or classification task. In the case of supervised learning, $\rho_{nk}$ is replaced by the true class labels.
All terms in Eq.~\ref{eq:lb_underst} can easily be computed batch-wise. The training procedure of the proposed tVAE
is summarized in Algorithm~\ref{tVAE:summary}.
\begin{figure*}[t!]
\centering
 \begin{minipage}{0.9\textwidth}
 \begin{algorithm}[H]
\small
\caption{Training procedure}\label{tVAE:summary}
\begin{algorithmic}[1]
  \State Initialize $\boldsymbol{\theta}, \boldsymbol{\phi}$ and $\boldsymbol{\xi}$
  \For {$\mathrm{epoch}=1,\ldots,\mathrm{max\_number\_epochs}$}
  \State $J=0$
  \For {$n=1,\ldots,N$} \Comment{encoding, decoding}
    \State $\big\{\boldsymbol{\mu}_n^{(\boldsymbol{x}|\boldsymbol{o})}, \log \boldsymbol{\sigma}_n^{(\boldsymbol{\boldsymbol{x}}|\boldsymbol{o})}\big\} 
		    = \texttt{Encoder}_{\boldsymbol{\phi}}(\boldsymbol{o}_n)$
    \State $\boldsymbol{\Sigma}^{(\boldsymbol{x}|\boldsymbol{o})}_n = \text{diag} \big\{ \big(\boldsymbol{\sigma}^{(\boldsymbol{x}|\boldsymbol{o})}_n\big)^2\big\}$
  \For {$t=1,\ldots,T$}
    \State $\boldsymbol{\epsilon}_t \sim \mathcal{N}\big(\boldsymbol{0}, \boldsymbol{I}\big)$
    \State $\boldsymbol{x}_{n,t} = \boldsymbol{\mu}_{n}^{(\boldsymbol{\boldsymbol{x}}|\boldsymbol{o})} 
		      + \boldsymbol{\sigma}_{n}^{(\boldsymbol{\boldsymbol{x}}|\boldsymbol{o})} \odot \boldsymbol{\epsilon}_t$
    \State $\big\{ \boldsymbol{\mu}_{n,t}^{(\boldsymbol{o}|\boldsymbol{x})} ,  \log \boldsymbol{\sigma}_{n,t}^{(\boldsymbol{o}|\boldsymbol{x})}\big\}
			= \texttt{Decoder}_{\boldsymbol{\theta}}(\boldsymbol{x}_{n,t})$
    \State $\boldsymbol{\Sigma}^{(\boldsymbol{o}|\boldsymbol{x})}_{n,t} = \text{diag} \big\{ \big(\boldsymbol{\sigma}^{(\boldsymbol{o}|\boldsymbol{x})}_{n,t}\big)^2\big\}$
    \EndFor
    \For {$k=1,\ldots,K$} \Comment{hyper-parameters}
      \State  $\alpha_k = \tfrac{\nu_k+D}{2}$
      \State $\beta_{nk} = \tfrac{1}{2}\big[ \nu_k +  \mathrm{Tr} \big\{\boldsymbol{\Sigma}^{(\boldsymbol{x}|\boldsymbol{o})}_n\boldsymbol{\Sigma}_k^{-1} \big\} + (\boldsymbol{\mu}^{(\boldsymbol{x}|\boldsymbol{o})}_n - \boldsymbol{\mu}_k)^T 
	      \boldsymbol{\Sigma}_k^{-1} (\boldsymbol{\mu}^{(\boldsymbol{x}|\boldsymbol{o})}_n - \boldsymbol{\mu}_k) \big]$
    \State $\ln qz_{nk} = \ln \pi_k + \tfrac{\nu_k}{2} \ln \tfrac{\nu_k}{2} - \ln \Gamma(\tfrac{\nu_k}{2}) - \tfrac{1}{2} \ln \det(\boldsymbol{\Sigma}_k) + \ln \Gamma\big(\alpha_k \big) -\alpha_k \ln \beta_{nk} $
    \State  $	\ln \rho_{nk} =  \ln qz_{nk} -H\big(\mathcal{G}(u_{nk}|\alpha_k, \beta_{nk})\big)
                    -\tfrac{D}{2}\ln\big\{2 \pi \big\} $
    \EndFor
    \State $\ln \boldsymbol{qz}_{n} = \big[\ln qz_{n1} , \ldots , \ln qz_{nK}\big]^T$
    \State $\boldsymbol{\gamma}_n = \texttt{Softmax}\big( \ln \boldsymbol{qz}_{n} \big)$
    \State $\boldsymbol{\gamma}_n = \big[\gamma_{n1} , \ldots , \gamma_{nK}\big]^T$
  \For {$k=1,\ldots,K$} \Comment{loss function}
  \State $J -= \gamma_{nk} ~ \ln \rho_{nk}$
    \EndFor
  \For {$t=1,\ldots,T$}
    \State $J -= \frac{1}{T}  \ln \mathcal{N}(\boldsymbol{o}_n |\boldsymbol{\mu}_{n,t}^{(\boldsymbol{o}|\boldsymbol{x})},\boldsymbol{\Sigma}_{n,t}^{(\boldsymbol{o}|\boldsymbol{x})})$
    \EndFor
    \State $J -=  H\big(\mathcal{N}(\boldsymbol{x}_n |\boldsymbol{\mu}_n^{(\boldsymbol{x}|\boldsymbol{o})}, \boldsymbol{\Sigma}_n^{(\boldsymbol{x}|\boldsymbol{o})} )\big) $
  \EndFor
    \State $J /= N$
    \State Determine gradients of parameters in 
	  $\boldsymbol{\theta}, \boldsymbol{\phi}$, $\boldsymbol{\xi}$
    \State Update parameters in $\boldsymbol{\theta}, \boldsymbol{\phi}$, $\boldsymbol{\xi}$
\EndFor
  \end{algorithmic}
\end{algorithm}
\end{minipage}
\end{figure*}

%% file: results.tex
The following section provides two experiments: Firstly, we will present a synthetic data experiment to demonstrate the properties of the \gls{tVAE} in an unsupervised scenario.
Secondly, we will consider authorship attribution, where we compare the proposed \gls{tVAE} algorithm with a \gls{gVAE} and (non)-linear \gls{SVM}. The \gls{gVAE} system was inspired by~\cite{Ebbers2017HiddenMM} and is, in structure, very similar to the method presented by~\cite{china}. For the sake of a fair comparison, we ensured that the network architecture of both, the tVAE and the gVAE implementations, were exactly the same\footnote{Except, of course, for the numeric differences in all trainable parameters, i.e. the neural network parameters as well as the mixture model parameters.}. Both algorithms are implemented in Python, where the training of the neural networks is accomplished via Tensorflow. The code is available at \url{https://github.com/boenninghoff/tVAE}.

Our implementations contains the following modifications relative to~\cite{Ebbers2017HiddenMM}: 
It must be ensured that the mixing weights sum to one and that each covariance matrix is invertible. Hence, instead of directly updating $\pi_k$, 
$\boldsymbol{\Sigma}_k$ and $\nu_k$, we introduced auxiliary variables such that
$\big[ \pi_1, \ldots , \pi_k\big]^T = \text{Softmax}\big(\big[m_1, \ldots , m_K \big]^T \big)$
,
$\nu_k = \log\big(\exp(n_k) + \exp(2.0 + \epsilon)\big)$
and 
$\boldsymbol{\Sigma}_k =  \boldsymbol{C}_k \boldsymbol{C}_k^T + \sigma_k^2 \boldsymbol{I}_{D \times D}$
where $\sigma_k^2$ is a fixed hyper-parameter~\cite{Ebbers2017HiddenMM}. 
More precisely, we enforce $\nu_k > 2$ by applying a modified Softplus-function and we enforce the positive definiteness of $\boldsymbol{\Sigma}_k$ through a Cholesky decomposition by constructing trainable lower-triangular matrices $\boldsymbol{C}_k$ with exponentiated (positive) diagonal elements.
We computed and updated the gradients of $n_k$, $m_k$ and $\boldsymbol{C}_k$ with respect to the loss function.

\subsection{Synthetic Data Experiment for Clustering}
\begin{figure*}[t]
\centering
\begin{subfigure}[t]{0.245\textwidth}
   \centering
  \includegraphics{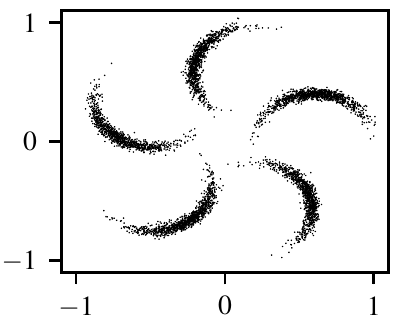}
  \caption{\noindent \small Training data points}
  \label{fig:synthetic_a}
\end{subfigure}
\hfill
\begin{subfigure}[t]{.245\textwidth}
  \centering
  \includegraphics{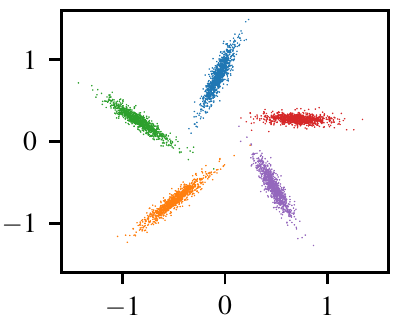}
  \caption{\noindent \small $\boldsymbol{x}_n \sim \mathcal{N}(\boldsymbol{\mu}_k, \boldsymbol{\Sigma}_k / u_{nk})$}
  \label{fig:synthetic_b}
\end{subfigure}
\hfill
\begin{subfigure}[t]{0.245\textwidth}
  \centering
  \includegraphics{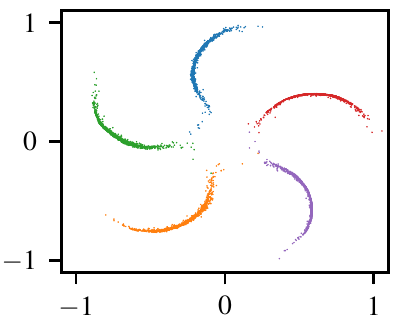}
  \caption{\noindent \small mean values $\boldsymbol{\mu}_{n}^{(\boldsymbol{o}|\boldsymbol{x})}$}
  \label{fig:synthetic_c}
\end{subfigure}
\hfill
\begin{subfigure}[t]{.245\textwidth}
  \centering
  \includegraphics{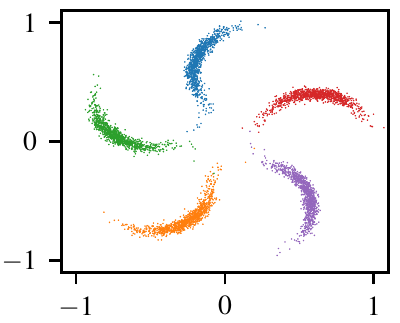}
  \caption{\noindent \small $\boldsymbol{o}_n \sim \mathcal{N} \big( \boldsymbol{\mu}_{n}^{(\boldsymbol{o}|\boldsymbol{x})}, \boldsymbol{\Sigma}_{n}^{(\boldsymbol{o}|\boldsymbol{x})} \big)$}
  \label{fig:synthetic_d}
\end{subfigure}
\vspace*{-0.3cm}
\caption{Results of the Student-$t$ autoencoder after training with the synthetic pinwheel data set.}
\label{fig:synthetic}
\vspace*{-0.4cm}
\end{figure*}

To demonstrate the properties of the tVAE, we first performed clustering on low-dimensional synthetic spirals of noisy data. 
The dataset as well as the clustering results are shown in Fig.~\ref{fig:synthetic}.
It is the same dataset used in~\cite{DilokthanakulMG16} to reproduce the illustrative results in~\cite{svae}. 
Encoder and decoder are fully connected feed-forward networks (with \texttt{ReLU} activation functions) of the form $L$-$H$-$H$-$D$ and $D$-$H$-$H$-$L$, respectively, where $L=2$ defines the dimension of the observations, $H=512$ represents the number of hidden nodes and $D=2$ is the latent space dimension.
We utilized Adam optimizer~\cite{Adam} to calculate the gradients.

A crucial aspect for a VAE with embedded mixture models is the aforementioned over-regularization behavior of VAE-based models occurring at the beginning of the training phase. 
Following~\cite{Overpruning}, it is caused by the regularization term of the ELBO in Eq.~\ref{eq:vae_lb_interpret}.
Both, the prior distribution in Eq.~\ref{eq:vae_enc_1} and the posterior in Eq.~\ref{eq:posterior_VAE} can be decomposed into univariate distributions and therefore we can also 
decompose the \gls{KL} divergence,
\begin{align}
 \text{KL} \big(q_{\boldsymbol{\phi}}(\boldsymbol{x}_n|\boldsymbol{o}_n) || p(\boldsymbol{x}_n) \big)
  = \textstyle{\sum_{d=1}^D} \text{KL} \big(q_{\boldsymbol{\phi}}(x_{n,d}|\boldsymbol{o}_n) || p(x_{n,d}) \big),
 \nonumber
\end{align}
where $x_{n,d}$ is the $d$-th component of $\boldsymbol{x}_{n}$. As mentioned in~\cite{Overpruning}, the model has to minimize the \gls{KL} term en bloc and not component-wise. 
One obvious option for the model is to enforce a large number of components $x_{n,d}$ helping to minimize the \gls{KL} term, which means these components are (close to) zero.
Similarly, our model has to maximize the cross-entropy in Eq.~\ref{eq:lb_underst}, which includes maximizing the term
 \begin{align}
 \beta_{nk} &\approx \mathrm{Tr} \big\{\boldsymbol{\Sigma}^{(\boldsymbol{x}|\boldsymbol{o})}_n\boldsymbol{\Sigma}_k^{-1} \big\} 
	+ (\boldsymbol{\mu}^{(\boldsymbol{x}|\boldsymbol{o})}_n - \boldsymbol{\mu}_k)^T 
	      \boldsymbol{\Sigma}_k^{-1} (\boldsymbol{\mu}^{(\boldsymbol{x}|\boldsymbol{o})}_n - \boldsymbol{\mu}_k)
 \end{align}
in Algorithm~\ref{tVAE:summary}.
What can now happen is that the entries of the covariance matrices $\boldsymbol{\Sigma}_k$ are getting continuously smaller, except for one global class. This leads to the 
``anti-clustering behavior'' that was observed in~\cite{china}.

To handle the over-regularization problem, we first trained a GMM to initialize the parameters $\pi_k$, $\boldsymbol{\mu}_k$ and $\boldsymbol{\Sigma}_k$ of each \mbox{Student-$t$} mixture component. A similar strategy was suggested in~\cite{DilokthanakulMG16}.
We found a simple modification to avoid merging all classes at the beginning: If we treat the obtained GMM weights as class labels for the first $10-15$ iterations (alternatively, one can randomly assign cluster labels), the neural networks become sufficiently stable and the merging effect is eliminated.
The degrees of freedom $\nu_k$ were initialized with $\approx 5$ for all $k$. The number of clusters $K$ was known \emph{a priori} and therefore kept fixed for all presented experiments.

After the completion of the training we used the decoder to sample new observations. Fig.~\ref{fig:synthetic_b} illustrates the linearly separable, learned manifolds in latent space by plotting samples drawn from each mixture component 
according to Eqs.~(\ref{eq:gp_u}) and~(\ref{eq:gp_x}). 
In Fig.~\ref{fig:synthetic_c} and~\ref{fig:synthetic_d} the mean values and new sampled observations are shown after applying the decoder to the latent 
data.

\subsection{Authorship Attribution}
In  the  previous  section,  we  have  shown  that  the Student-$t$ model has the ability to learn a 
non-linear generative process. Next,  we  examine  the  influence  of  the $\nu_k$-values by comparing the Student-$t$ model with a Gaussian model for the authorship attribution task on Amazon reviews.

\subsubsection{Feature Extraction}
Our \textsc{AdHominem} system originally addresses the authorship verification task: Given two documents, decide, whether they were written by the same person or not.
The purpose of \textsc{AdHominem} here is to provide a feature extraction module, in which we encode the stylistic characteristics of a document $\mathcal{D}$ of variable length consisting of characters, words, and sentences into a single fixed-length \textit{linguistic embedding vector} (LEV), denoted by 
$\boldsymbol{o}_n = \mathcal{A}(\mathcal{D}_n)$ for the $n$-th observed LEV.
The core of \textsc{AdHominem} is a two-level hierarchical attention-based bidirectional LSTM network~\cite{hochreiter1997long}. Besides pre-trained word embeddings, \textsc{AdHominem} also provides a characters-to-word encoding layer to take the specific uses of prefixes and suffixes as well as spelling errors into account. A detailed description can be found in ~\cite{BigData19}.

\subsubsection{Amazon Reviews Dataset}
\textsc{AdHominem} was trained on a large-scale corpus of short Amazon reviews. The dataset is described in~\cite{BigData19} and consists of $9,052,606$ reviews written by $784,649$ authors, with document lengths varying between $80$ and $1000$ tokens.
In this work, we randomly selected $21,172$ reviews written by 30 authors, which were not involved in the training procedure of \textsc{AdHominem}. Each author contributes with at least $503$ reviews and with a maximum of $1,000$ reviews.

\subsubsection{Hyper-Parameter Tuning and Regularization}
Encoder and decoder were fully connected feed-forward networks (with \texttt{tanh} activation function) of the form $L$-$H$-$D$ and $D$-$H$-$L$, respectively, where $L=200$ defines the dimension of the observations, $H=\frac{D+L}{2}$ represents the number of hidden nodes and $D$ is the latent space dimension.
In all experiments, the Adam optimizer proposed in~\cite{Adam} was used to update the model parameters. Gradients are normalized so that their $l_2$-norm is less than or equal to $1$.
Furthermore, we add an $l_1$-regularization term,
\begin{align}
J_{\boldsymbol{\phi},\boldsymbol{\theta}}
 = \beta \cdot \sum_{\forall (\boldsymbol{W},\boldsymbol{b})   \in \{\boldsymbol{\phi},\boldsymbol{\theta}\}}  
        \norm{\text{vec}(\boldsymbol{W})}_1 + \norm{\boldsymbol{b}}_1 ,
\end{align}
to reduce overfitting. The terms $\boldsymbol{W},\boldsymbol{b}$ represent the weights and bias terms of the encoder/decoder networks.
Our hyper-parameter tuning is based on a grid search over the following parameter-set combinations:
\begin{align*}
    \text{stepsize } \alpha &\in \{0.001, 0.002, 0.003, 0.004, 0.005\}, \\
    \sigma_k^2 &\in \{0.001, 0.01, 0.1, 0.5, 0.9\}, \\
    D &\in \{20, 50, 100, 150, 200\}, \\
    \beta &\in \{0.001, 0.005, 0.01, 0.05\}. 
\end{align*}

\subsubsection{Results}
To evaluate the models in terms of the average error rates, we performed a $5$-fold cross-validation.  
In a first step, we held out 20\% of the available data, which was equally split into a development set and a test set.
Additionally, we addressed the challenges of training the autoencoders with a small number of labeled data items by varying the size of the labeled data from 20\% to 100\% (reviews were dropped out randomly).  
The remaining document pairs were used to fit the model parameters.
Using the best models found (depending on the hyper-parameters) we evaluated the performance of all methods with respect to the average error rate.

\begin{figure}[tb]
	\centering
	\includegraphics[width=.5\textwidth]{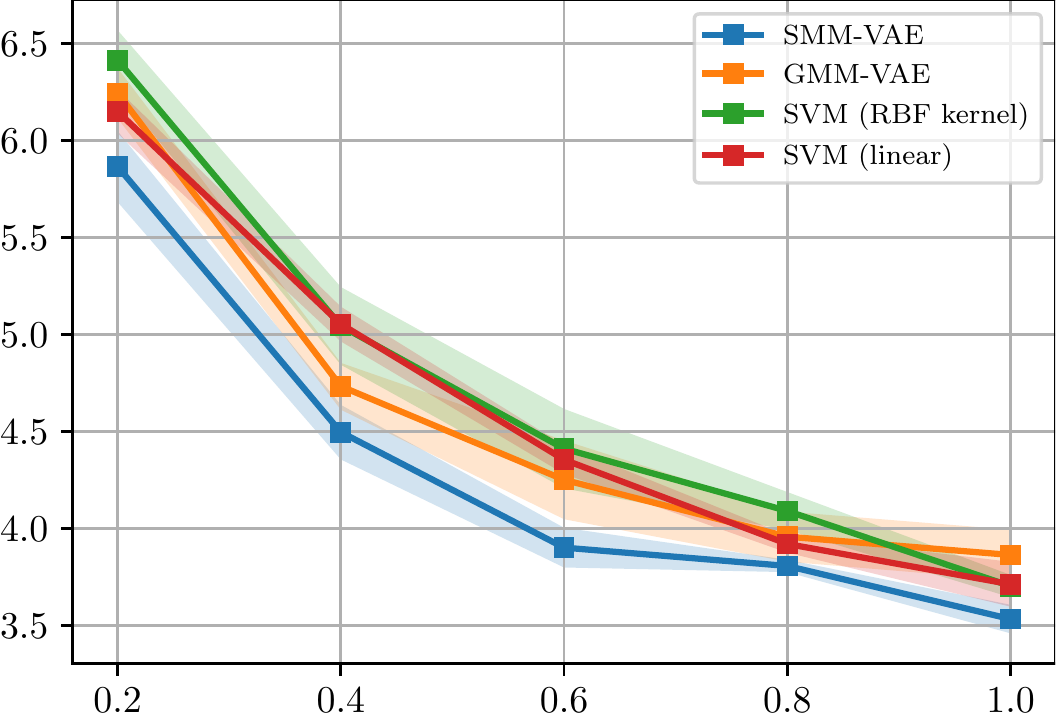}
	\caption{Average error rates for supervised authorship attribution over a $5$-fold cross-validation with varied number of labeled training pairs (from 20\% to 100\%).}
	\label{fig:re1}
\end{figure}

\begin{figure}[t!]
	\centering
	\begin{minipage}[b]{.445\textwidth}
	\includegraphics[width=.95\textwidth]{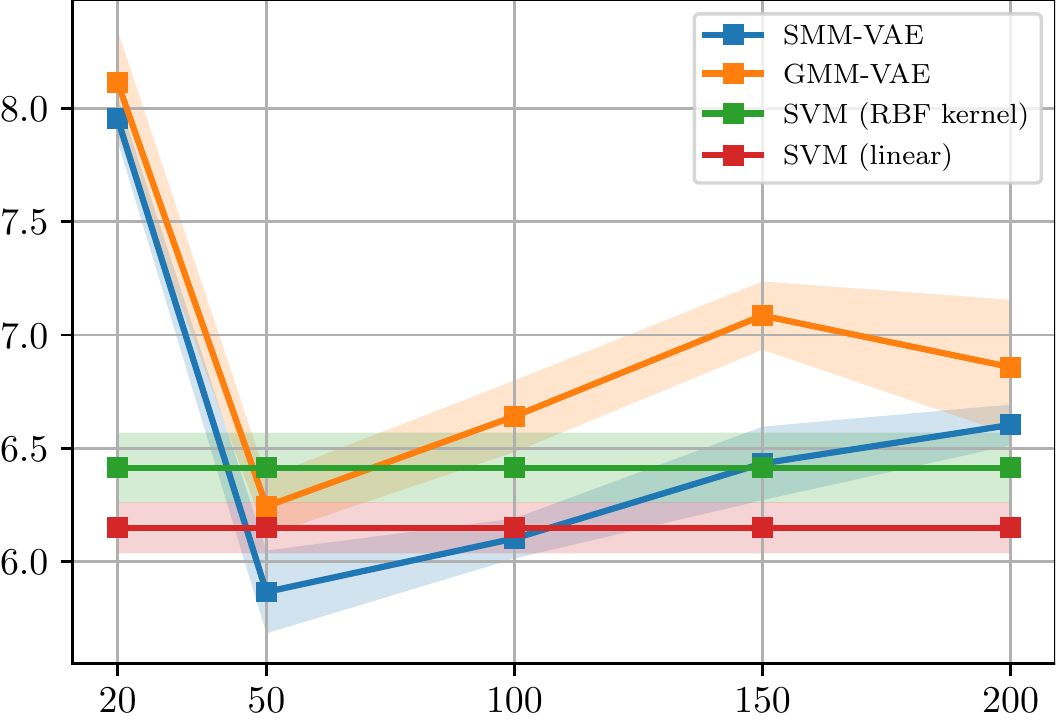}
	\centerline{(a) 20\% of training data} \medskip
    \end{minipage}
    \hspace{0.4pt}
    \begin{minipage}[b]{.445\textwidth}
    \includegraphics[width=.95\textwidth]{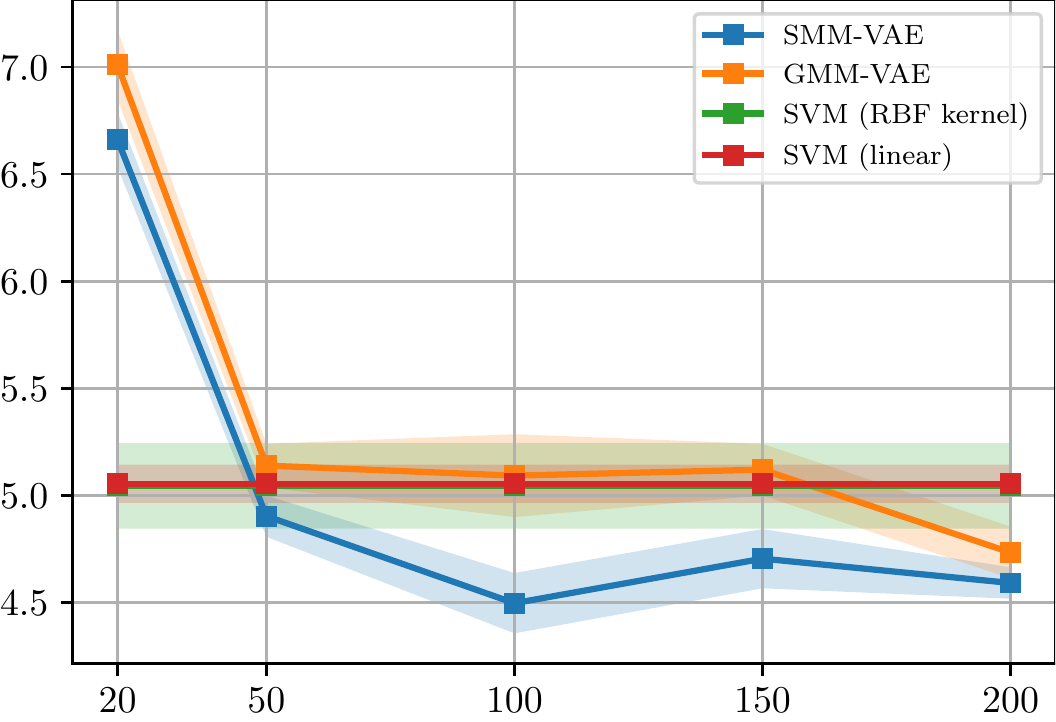}
	\centerline{(b) 40\% of training data} \medskip
    \end{minipage}
    \begin{minipage}[b]{.445\linewidth}
    \includegraphics[width=.95\textwidth]{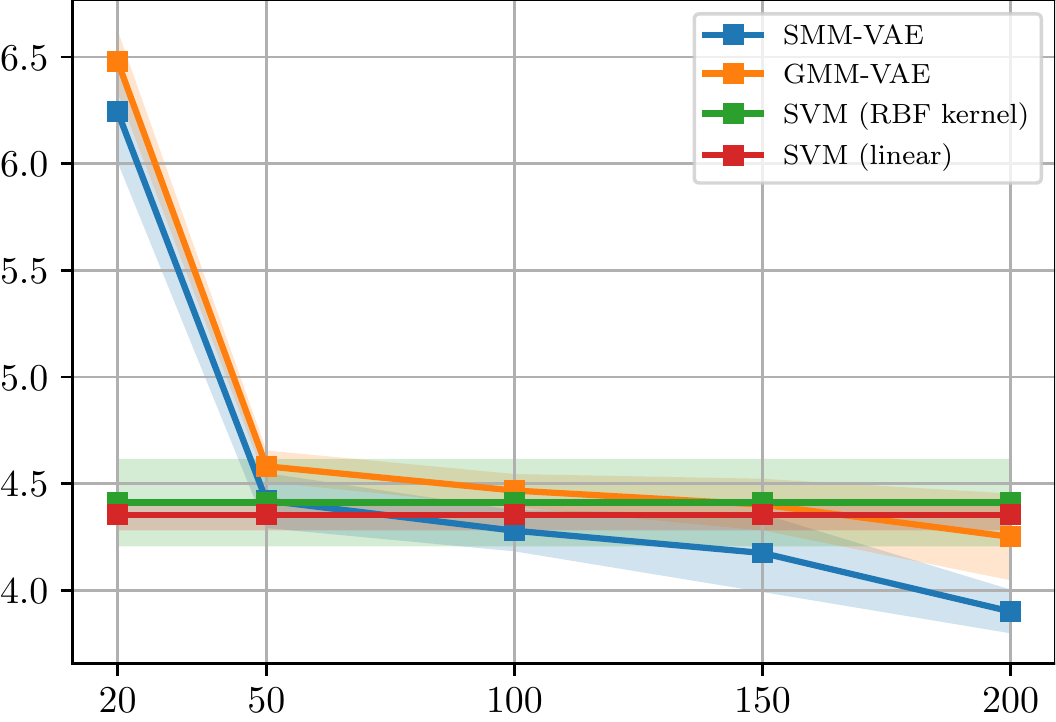}
	\centerline{(c) 60\% of training data}\medskip
    \end{minipage}
	\hspace{0.4pt}
    \begin{minipage}[b]{.445\linewidth}
    \includegraphics[width=.95\textwidth]{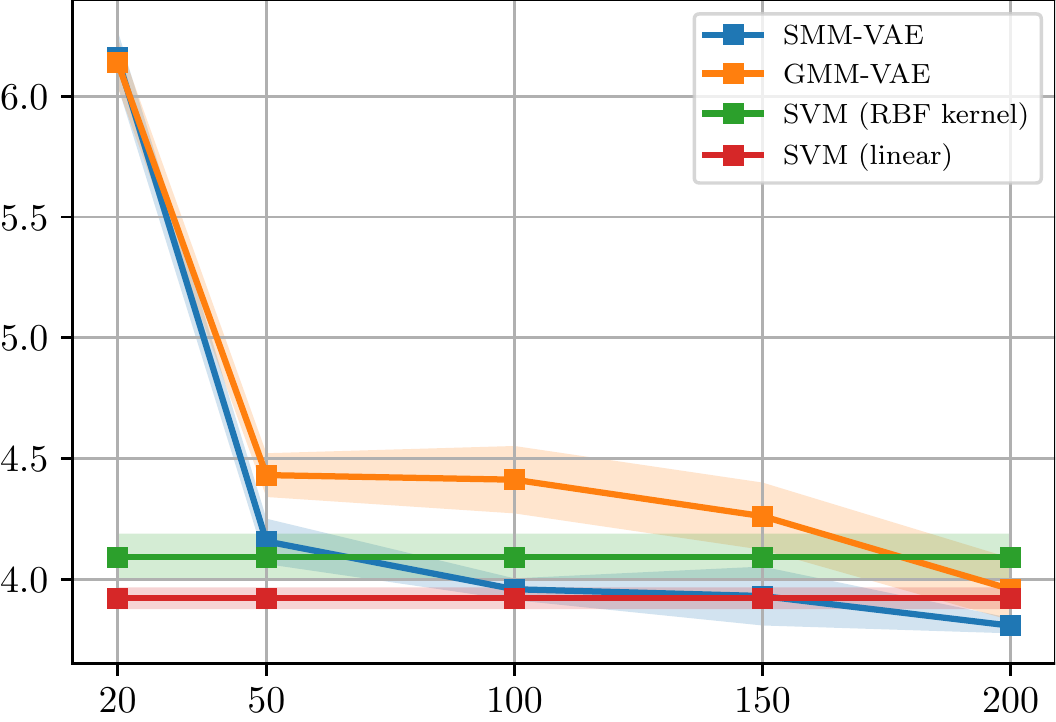}
    \centerline{(d) 80\% of training data}\medskip
    \end{minipage}
    \begin{minipage}[b]{.445\linewidth}
    \includegraphics[width=.95\textwidth]{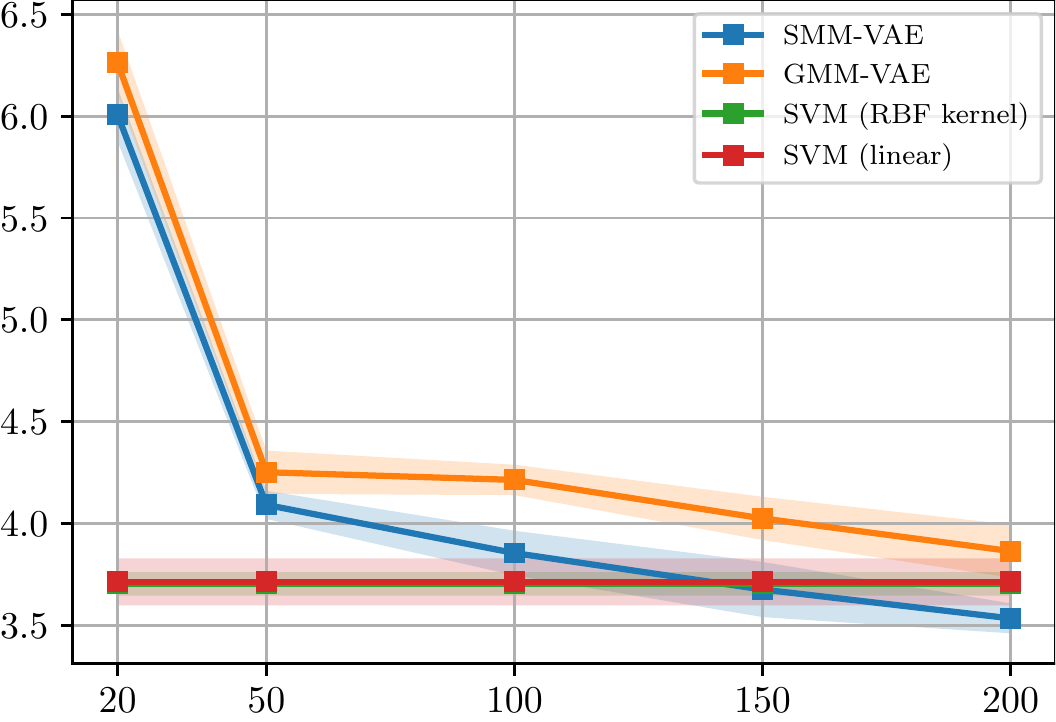}
    \centerline{(e) 100\% of training data}\medskip
    \end{minipage}
	
	\caption{Average error rates with a varied size of the dimension of the latent variable (from $D=20$ to $D=200=L$).}
	\label{fig:re2}
\end{figure}

\begin{table}[t!]
\caption{Average error rates and standard deviations for supervised authorship attribution over a $5$-fold cross-validation.}
\vspace*{0.3cm}
\begin{tabular}{r||c|c|c|c} 
        size of training set
            &\textsc{SMM-VAE}
            &\textsc{GMM-VAE}
            &\text{SVM (linear)}
            &\text{SVM (RBF kernel)}
            \\ \hline\hline
    ~$20\%$ ($4,234$ texts)   
        &$\boldsymbol{5.86\pm 0.60}$
        &$6.24 \pm 0.43$ 
        &$6.15 \pm 0.37$ 
        &$6.41 \pm 0.51$ 
        \\  \hline
    ~$40\%$ ($8,468$ texts)  
        &$\boldsymbol{4.49\pm 0.46}$
        &$4.73 \pm 0.40$ 
        &$5.05 \pm 0.29$ 
        &$5.04 \pm 0.66$
        \\  \hline
    ~$60\%$ ($12,703$ texts)  
        &$\boldsymbol{3.90\pm 0.34}$
        &$4.25 \pm 0.67$
        &$4.35 \pm 0.24$
        &$4.41 \pm 0.68$
        \\  \hline
    ~$80\%$ ($16,937$ texts)  
        &$\boldsymbol{3.80\pm 0.10}$
        &$3.95 \pm 0.43$
        &$3.92 \pm 0.14$
        &$4.09 \pm 0.32$
        \\  \hline
    ~$100\%$ ($21,172$ texts)   
            &$\boldsymbol{3.53 \pm 0.24}$
         &$3.86 \pm 0.43$
         &$3.71 \pm 0.38$
         &$3.70 \pm 0.19$
  \end{tabular}
\label{tab:acc1}
\end{table}
In Fig~\ref{fig:re1} and Table~\ref{tab:acc1} we summarize the classification error rates using our approach versus the Gaussian VAE as well as (non)-linear SVMs. The lowest error rates for each setup are displayed in bold face in Table~\ref{tab:acc1}.
It can be seen that our Student-$t$ model is able to (slightly) outperform all baseline methods.
For all methods, the performance gradually improves as the number of training pairs is increased.
In addition, Fig.~\ref{fig:re2} shows the performance results w.r.t.~the dimension of the latent variable, where we can
make the following observation: the best choice for $D$ increases when more reviews are added to the training set. 
For $20\%$ of the training data, the optimal dimension is $D=50$, for $40\%$ we have $D=100$ and for more than $60\%$, $D=200$ yields the lowest error rate.

\begin{figure}[t!]
	\centering
	\begin{minipage}[b]{.445\textwidth}
	\includegraphics[width=.95\textwidth]{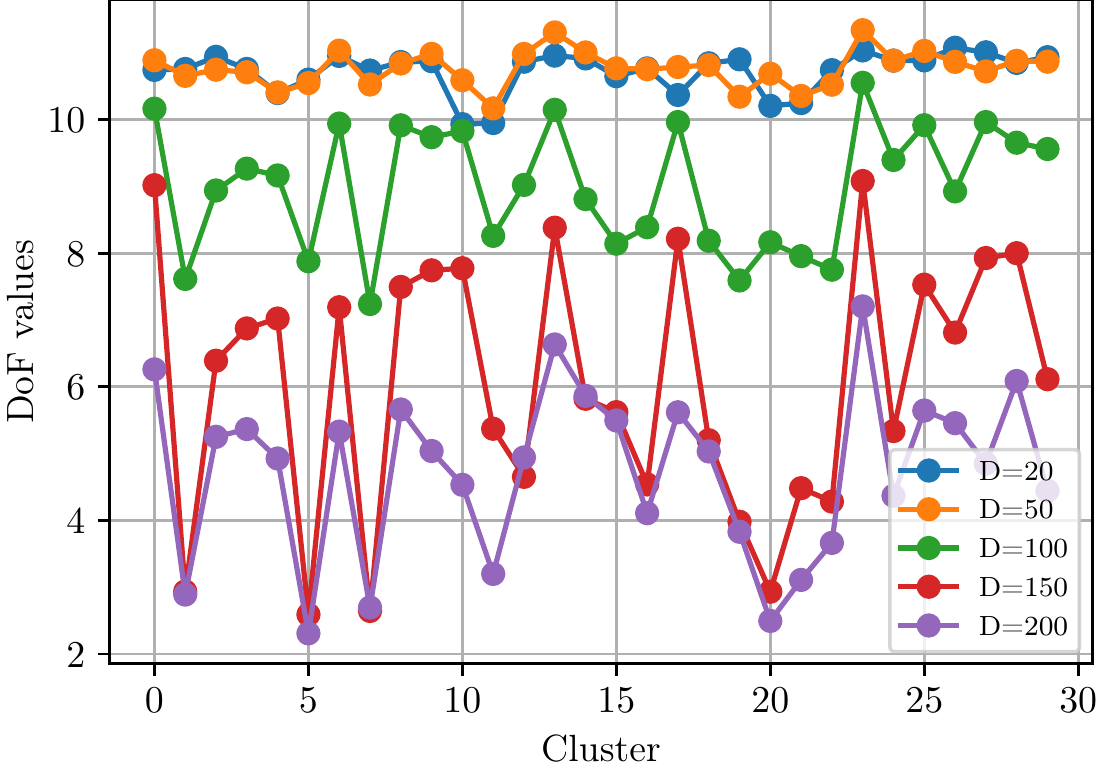}
	\centerline{(a) 20\% of training data} \medskip
    \end{minipage}
    \hspace{0.4pt}
    \begin{minipage}[b]{.445\textwidth}
    \includegraphics[width=.95\textwidth]{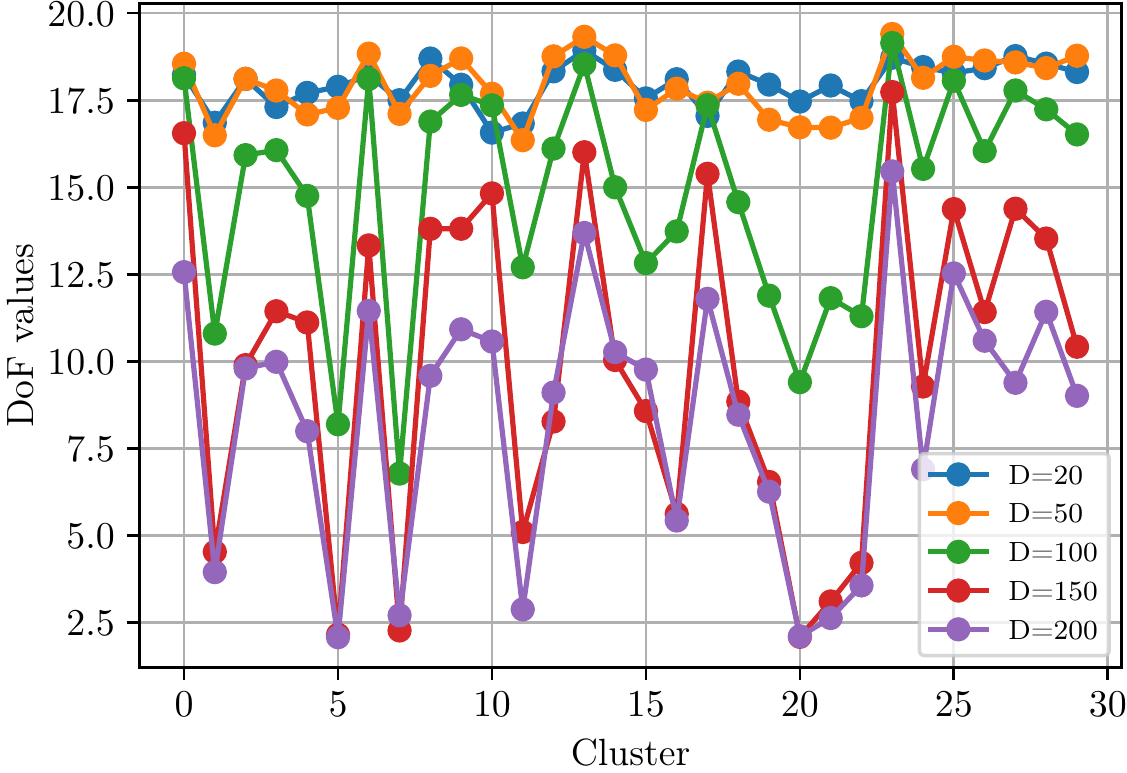}
	\centerline{(b) 40\% of training data} \medskip
    \end{minipage}
    \begin{minipage}[b]{.445\linewidth}
    \includegraphics[width=.95\textwidth]{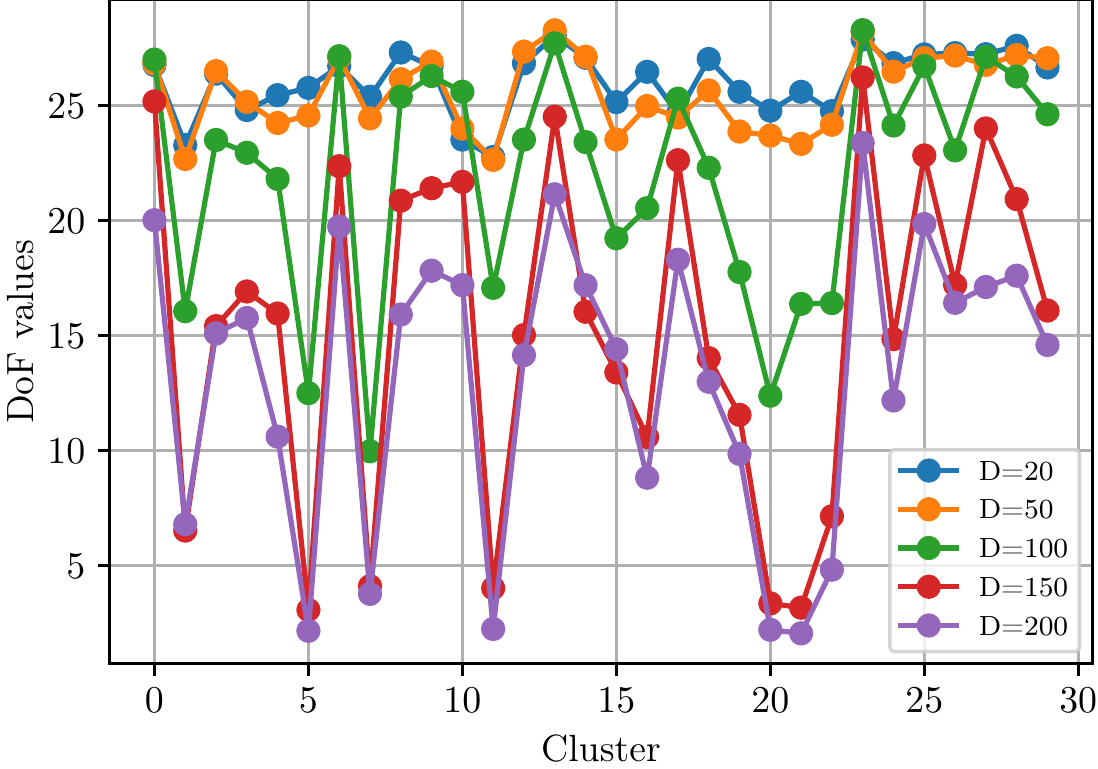}
	\centerline{(c) 60\% of training data}\medskip
    \end{minipage}
	\hspace{0.4pt}
    \begin{minipage}[b]{.445\linewidth}
    \includegraphics[width=.95\textwidth]{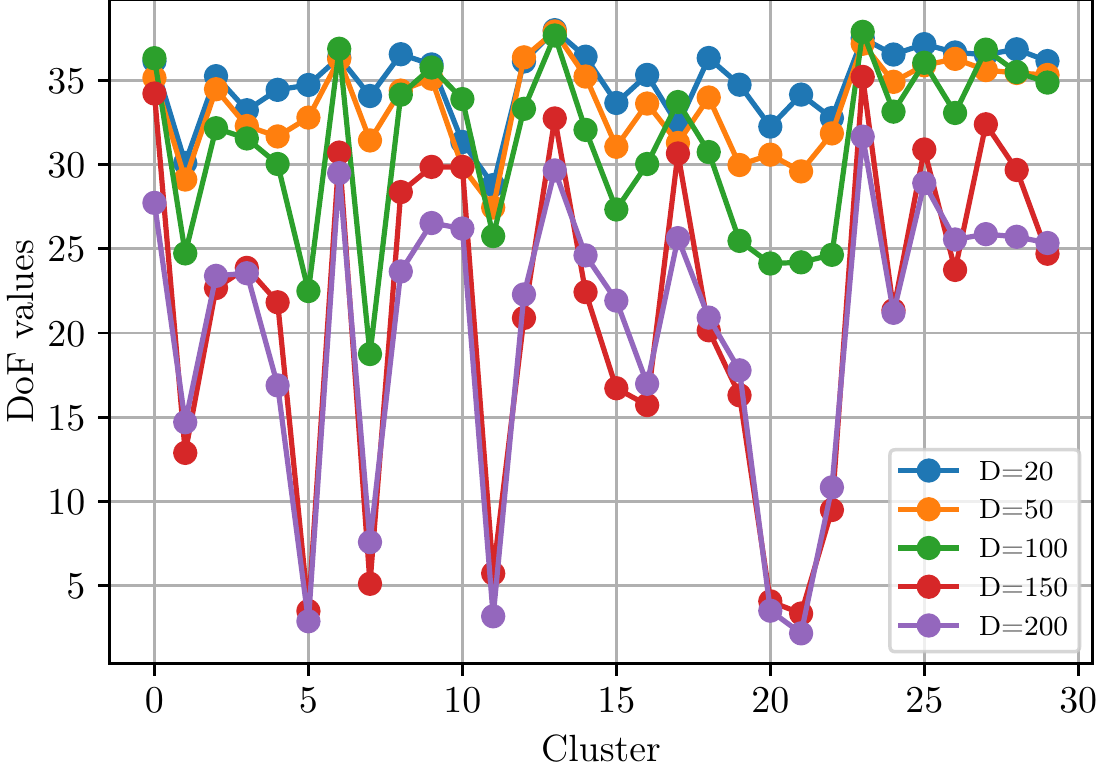}
    \centerline{(d) 80\% of training data}\medskip
    \end{minipage}
    \begin{minipage}[b]{.445\linewidth}
    \includegraphics[width=.95\textwidth]{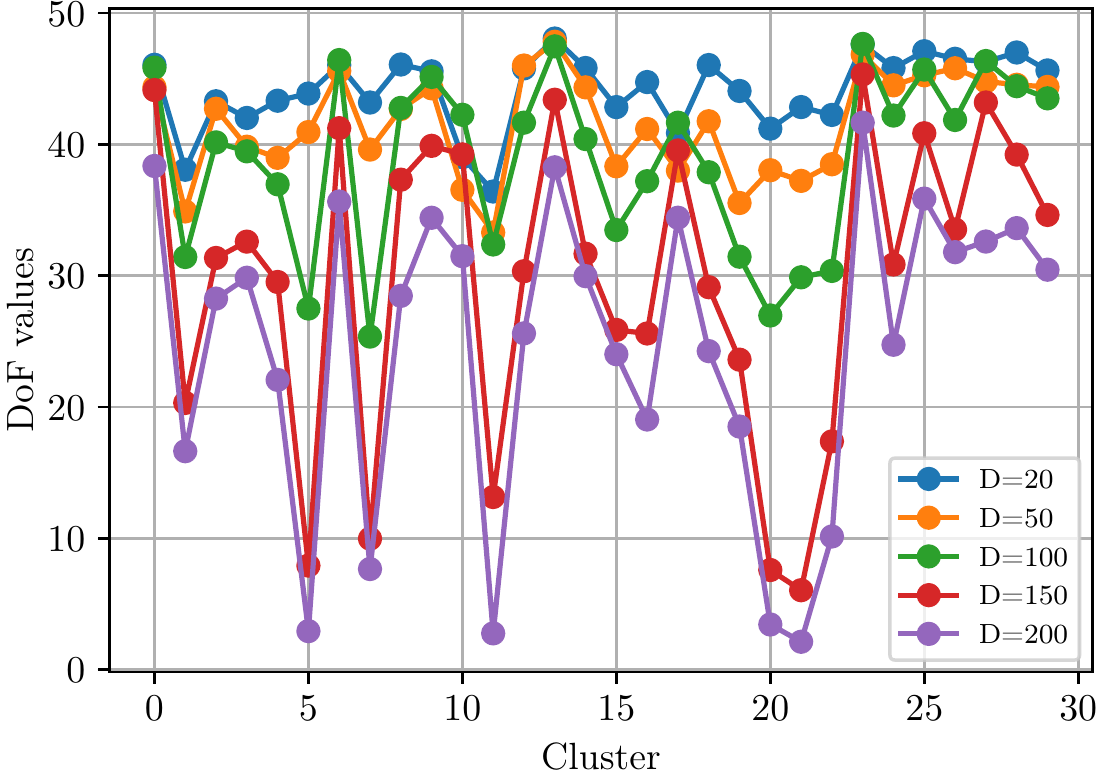}
    \centerline{(e) 100\% of training data}\medskip
    \end{minipage}
	\caption{Degrees of freedom for all clusters (i.e.~authors) w.r.t. the latent space dimension $D$.}
	\label{fig:re3}
\end{figure}

Fig.~\ref{fig:re3} presents the learned degrees of freedom ($\nu_k$) for all clusters (i.e.~authors). Again, the latent space dimension and the size of the training set were varied as described above. The plots clearly show that for lower dimensions, the cluster distributions are approximately Gaussian. With the increase of the latent dimension $D$, the mixture components become more heavy-tailed, making the Student-$t$ distribution a better fit. 

%% file: conclusion.tex
Variational autoencoders have proven their benefit in many tasks, while providing an attractive machine-learning framework that combines many strengths of neural-network training with the uncertainty metrics of statistical models. They can learn a low-dimensional manifold to summarize the most salient characteristics of data and come with a natural, statistical interpretation, both in the latent as well as in the observation space.
In our work, we are addressing the question of whether they can benefit from a distributional model that allows for more heavy-tailed distributions, with the intuition of limiting susceptibility to outliers in the latent space and of improving modeling capacity.

Towards that goal, we have proposed and evaluated a VAE that is equipped with an embedded Student-t mixture model. It incorporates an assumption of 
Student-t distributed data into the joint learning mechanism for the latent manifold and its statistical distribution.
Variational inference is performed by trying to simultaneously solve both tasks: jointly learning a nonlinear mapping to transform a given dataset of interest onto a 
(lower-dimensional) manifold in latent space {\em and\/} grouping the latent representations into meaningful categories.

We have derived a variational learning algorithm to accomplish this goal and we have shown its benefit for learning latent representations, both on toy data as well as on the real-world task of authorship attribution, where the more flexible model has proven its capability to obtain better results than both SVM-based classifiers as well as the standard Gaussian VAE.